\documentclass[10pt,twocolumn,letterpaper]{article}

\usepackage{cvpr}              
\usepackage{graphicx} 
\usepackage{multirow}
\usepackage{xcolor}
\usepackage{amsmath}
\usepackage{graphicx}
\usepackage{booktabs}
\usepackage{bbding}
\usepackage{float}
\usepackage{wrapfig}
\definecolor{dgreen}{rgb}{0.0,0.6,0.0}
\usepackage{colortbl}  
\usepackage{xcolor}
\usepackage{array}

\usepackage{booktabs}
\usepackage{longtable}
\usepackage{tabularx} 
\usepackage{booktabs} 
\usepackage{ragged2e} 
\usepackage{xargs}
\usepackage[colorinlistoftodos,prependcaption,textsize=normalsize]{todonotes}
\newcommandx{\xiangyu}[2][1=]{\todo[linecolor=red,backgroundcolor=red!25,bordercolor=red,#1]{#2}}

\marginparwidth=\dimexpr \marginparwidth + 1.5cm\relax

\definecolor{cvprblue}{rgb}{0.21,0.49,0.74}
\usepackage[pagebackref,breaklinks,colorlinks,allcolors=cvprblue]{hyperref}

\def\paperID{8324} 
\def\confName{CVPR}
\def\confYear{2025}

\title{SemGeoMo: Dynamic Contextual Human Motion Generation \\with Semantic and Geometric Guidance}

\author{\textbf{  
Peishan Cong$^{1,2,}$\thanks{Equal contribution. $\dagger$ Corresponding author. This work was supported by NSFC (No.62206173, No. 62306261), Shanghai Frontiers Science Center of Human-centered Artificial Intelligence (ShangHAI), MoE Key Laboratory of Intelligent Perception and Human-Machine Collaboration (KLIP-HuMaCo), 
and the Shun Hing Institute of Advanced Engineering (SHIAE) No. 8115074.},
Ziyi Wang$^{1,}\footnote[1]{}$,
Yuexin Ma$^{1,}\footnote[2]{}$, Xiangyu Yue $^{2,}\footnote[2]{}$}\\ 
$^{1}$ ShanghaiTech University
$^{2}$ The Chinese University of Hong Kong \\
{\tt\small \{congpsh,wangzy17,mayuexin\}@shanghaitech.edu.cn}}

\begin{document}
\maketitle
\begin{abstract}

Generating reasonable and high-quality human interactive motions in a given dynamic environment is crucial for understanding, modeling, transferring, and applying human behaviors to both virtual and physical robots. In this paper, we introduce an effective method, SemGeoMo, for dynamic contextual human motion generation, which fully leverages the text-affordance-joint multi-level semantic and geometric guidance in the generation process, improving the semantic rationality and geometric correctness of generative motions. Our method achieves state-of-the-art performance on three datasets and demonstrates superior generalization capability for diverse interaction scenarios. The project page and code can be found at
\url{https://4dvlab.github.io/project_page/semgeomo/}.
\end{abstract}    
\vspace{-4ex}
\section{Introduction}
\label{sec:intro}
%


Dynamic contextual human motion generation~\cite{li2023object,jiang2024scaling,li2023controllable} aims to generate human interaction motions that are both commonsense and geometrically accurate, adapting seamlessly to real dynamic environments. Its core lies in constructing an interaction-oriented world model for humans, enabling reasonable adaptation to changes of interactive objects or people. This interaction-oriented world model can support a wide range of applications, including human-robot interaction, closed-loop simulators, intelligent sports coaching, and immersive VR/AR gaming experiences.


As the importance of interaction becomes increasingly recognized, some studies have evolved from text-driven human motion generation~\cite{mdm2022human,mdmprior,mld,motiondiffuse,Mofusion,jiang2024motiongpt} to text-driven joint generation of human-object or human-human interactions~\cite{li2023object,bhatnagar2022behave,liang2024intergen,xu2024inter,wang2023intercontrol}. However, generating motions jointly for both the human and the interactive target creates an excessively large search space, often leading to suboptimal generation results. Additionally, the lack of fine-grained control over the generated data hinders the creation of personalized interactive motions, limiting its applicability to real-world scenarios such as humanoid operation and human-robot interaction.

\begin{figure}[t]
\vspace{-2ex}
\centering\includegraphics[width=0.98\linewidth]{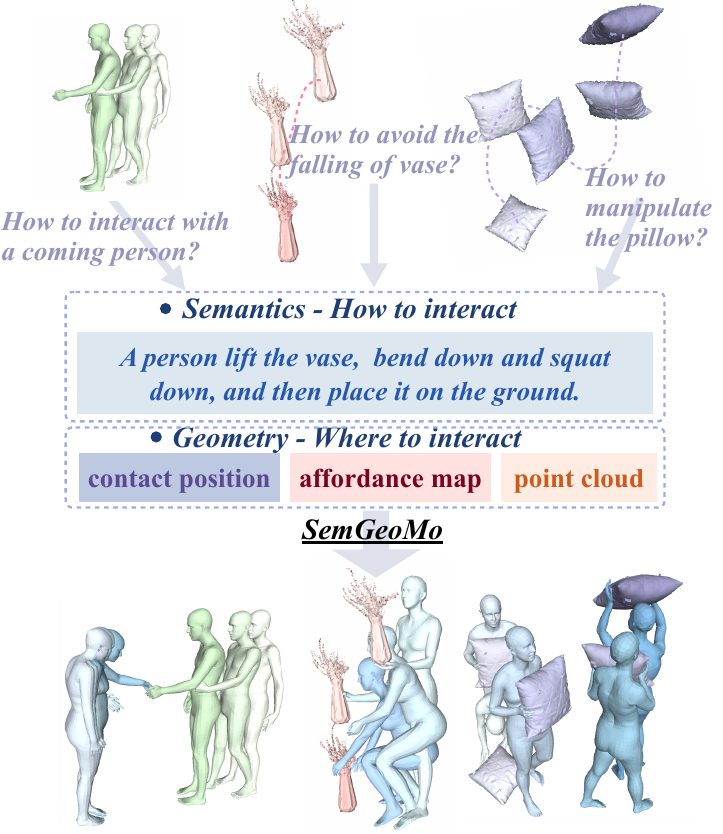}
    \caption{Given sequential point clouds of interactive targets, SemGeoMo generates realistic and high-quality human interactive motions along with corresponding textual descriptions. By leveraging both semantic and geometric guidance, our method ensures the semantic coherence and geometric accuracy of the generated results.}
    \label{fig:teasor}
    \vspace{-4ex}
\end{figure}
Recently, contextual human motion generation has garnered increasing attention for its ability to produce controllable interactions based on specified scenario conditions. However, many works~\cite{scenediff,wang2024move,cen2024generating,wang2021synthesizing,xiao2023unified} only focus on generating coarse-grained human trajectories and motions in static environments with fixed furniture or layouts rather than the fine-grained details of interactions, limiting adaptability to dynamic conditions. A few recent studies~\cite{cong2024laserhuman,li2023object,xu2024regennet} have begun to explore dynamic contextual human motion generation. However, these approaches have notable limitations: 1) they lack textual guidance, which undermines the semantic coherence of interactions and limits the generalization capability of the approach, and 2) they fail to incorporate fine-grained geometric representations, resulting in insufficient constraints on the geometric accuracy of generated interaction motions.


In this work, we propose a novel dynamic contextual human motion generation method, named \textbf{SemGeoMo}, illustrated in Fig.\ref{fig:teasor}, which could generate reasonable and high-quality interactive motions by comprehensively integrating semantic information from textual descriptions with hierarchical geometric features extracted from interactive objects. The first challenge is \textit{how to construct the semantic guidance}.
Given that large language models (LLMs) possess general knowledge and can provide rich information on the attributes of interacting objects as well as guidance for the interaction process, we introduce an automated interaction text annotator. By leveraging careful prompt design and fine-tuning of the model~\cite{roumeliotis2023chatgpt,touvron2023llama}, our \textbf{LLM Annotator} eliminates the need for manual text labeling, offering strong support for the semantic coherence and generalization of generated human interaction behaviors. 
The second challenge is \textit{how to construct the geometric guidance}. To ensure the geometric accuracy of interactive motions, such as avoiding geometric penetration and ensuring appropriate contact between human and object, we propose a two-stage framework that decouples contact geometry generation from interactive motion generation. 
In the first stage, \textbf{SemGeo Hierarchical Guidance Generation}, a diffusion model generates affordance-level and joint-level interaction cues guided by semantic information to capture both coarse and precise geometric positioning. 
In the second stage, \textbf{SemGeo-guided Motion Generation}, these cues are effectively utilized to guide the generation of detailed human motions, improving both semantic plausibility and geometric accuracy. 
It is worth noting that our model simultaneously generates human motions and language descriptions at varying levels of granularity, which not only improves the quality of the generated motions but also enhances the interpretability and comprehensibility of the interactions.
Extensive experiments demonstrate that we achieve state-of-the-art performance on three human-object interaction datasets.
Moreover, we demonstrate the generalization capability of our method in more challenging scenarios, including interactions with unseen objects, human-human interactions, and interactions with deformable objects. 
To summarize, our work makes the following contributions:
\begin{itemize}
\item We propose a novel method that generates responsive human motions and corresponding textual descriptions based on observed dynamic interactive targets. 

\item We introduce an automated text annotator that interprets reactions during interactions, reducing the burden of manual labeling and enhancing the generalization capability.

\item Our method fully utilizes multi-level semantic and geometric guidance—including text, affordance, and joint-level cues—throughout the generation process, improving both the semantic rationality and geometric accuracy of interactive motion.

\item Our method generates high-quality human motions and achieves state-of-the-art performance across three benchmarks and an unseen dataset.

\end{itemize}

\section{Related Work}
\begin{figure*}[ht!]
    \centering   \includegraphics[width=1.0\linewidth]{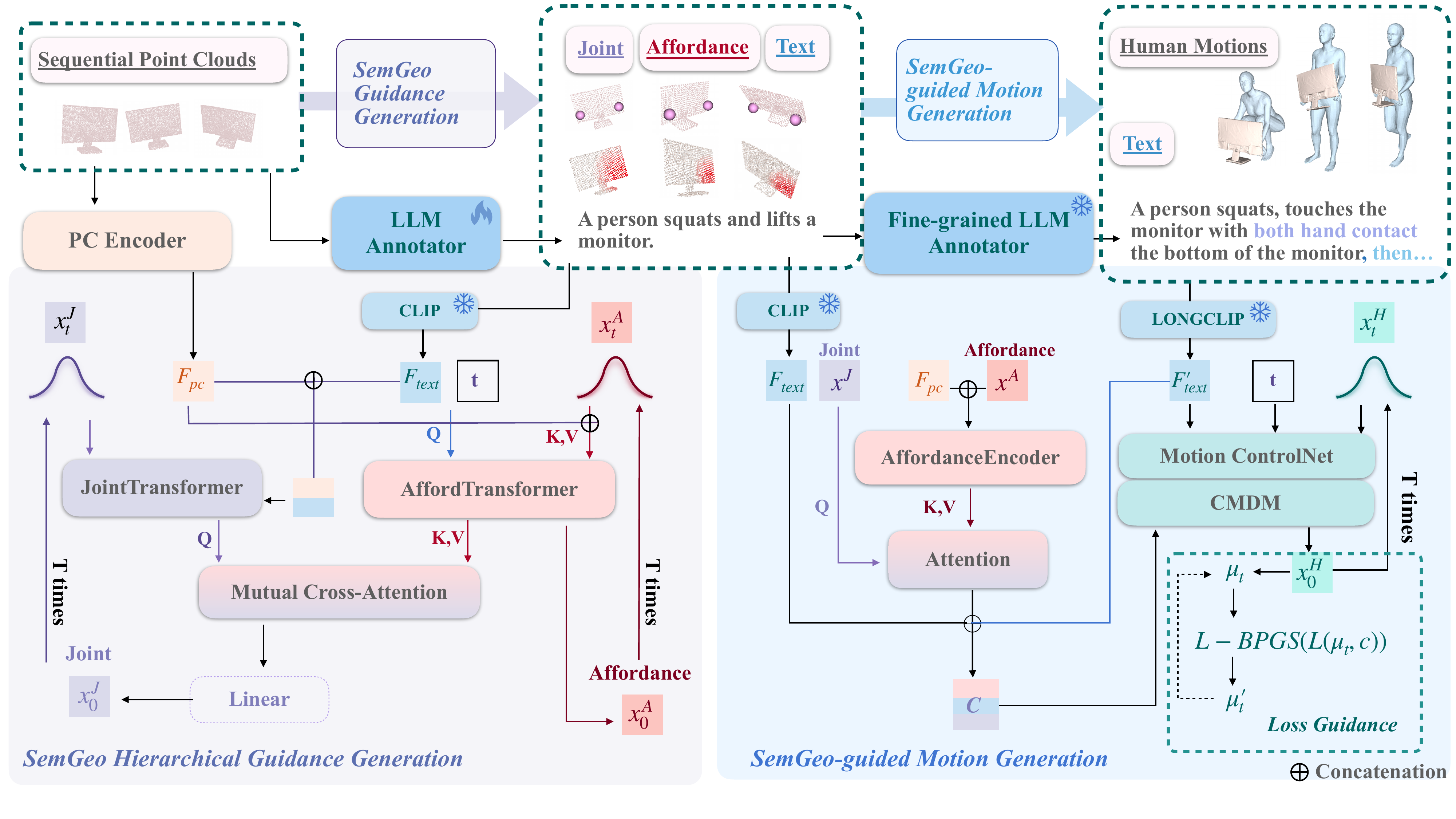}
     \vspace{-7ex}
    \caption{The pipeline of our two-stage framework. LLM Annotator provides the semantic guidance. SemGeo Hierarchical Guidance Generation takes textual information and sequential point cloud as condition and generate affordance-level and joint-level guidance. Then SemGeo-guided Motion Generation utlizes semantic and geometric information to generate responsive human motion.}
    \label{fig:framework}
    \vspace{-4ex}
\end{figure*}
\subsection{Text-guided Human Motion Generation}
Text-conditioned human motion generation~\cite{mdm2022human,mdmprior,motiondiffuse,Mofusion,jiang2023motiongpt,zhang2024motiondiffuse,dabral2023mofusion,wan2023tlcontrol} have made significant progress with the rise of diffusion models~\cite{ho2020denoising,ho2022classifier}. With the emergence of human-object interaction datasets~\cite{li2023object,bhatnagar2022behave,zhang2023neuraldome,Zhao_2024_CVPR} and human-human interaction datasets~\cite{liang2024intergen,xu2024inter}, some works have begun to explore text-driven interaction motion generation.
Several studies~\cite{peng2023hoi,diller2024cg,wu2024thor,xu2023interdiff} dive into jointly generating a sequence of human and object poses based on textual conditions.
HOI-Diff~\cite{peng2023hoi} emphasizes the significance of affordance information and Thor~\cite{wu2024thor} refines object rotation during each inverse diffusion step for human-object interaction. Furthermore, CHOIS~\cite{li2023controllable} further integrates the 2D waypoints with object geometry loss during sampling process. 
Other studies~\cite{xu2024inter,liang2024intergen} focus on human-human interaction, where two human motions are jointly generated based on language descriptions. 
However, the joint generation of human motions and interactive targets creates a large optimization search space during training, leading to lower quality in the generated motions. Moreover, conditioning solely on text lacks fine-grained control over the generation process, limiting its applicability in areas such as robotic operations, human-robot interaction, and AR/VR immersive experiences.


\subsection{Contextual Human Motion Generation}
Contextual human motion generation explores settings that are more applicable to real-world scenarios, where human motion is influenced by and interacts with the given environment.
Several works~\cite{kulkarni2024nifty,jiang2023full} focus on interactions with seating furniture~\cite{jiang2022chairs} and others~\cite{scenediff, wang2024move, cen2024generating, wang2021synthesizing} generate natural human motions in 3D indoor scenes~\cite{xiao2023unified,araujo2023circle, jiang2024scaling, wang2022humanise, hassan2019resolving,yi2025generating}. 
SceneDiff~\cite{scenediff} utilizes point clouds as conditions to generate feasible interactions. AffordMotion~\cite{wang2024move} provides a two-stage framework that employs a scene affordance map as an intermediary.
However, these works are limited to static environments, where interacted objects are often restricted to fixed furniture like beds and chairs, and the category of motions is limited to sitting, lying, and walking. They focus on the trajectory and the goal state, rather than interacting with dynamic, ever-changing targets.

Following works~\cite{li2023object,xu2024regennet, cong2024laserhuman} introduce dynamic, interactive targets, such as movable objects or other people. ReGenNet~\cite{xu2024regennet} 
generates human reactions conditioned on given human motion in SMPL~\cite{loper2023smpl} representation and the action condition. However, SMPL representation requires additional processing on raw sensor data and is not suitable for all dynamic targets such as objects. A more flexible point cloud-based representation is used in~\cite{cong2024laserhuman} for interacting scenes or objects. However, it lacks a carefully designed feature modeling method for geometric and temporal information, resulting in suboptimal performance.
OMOMO~\cite{li2023object} presents a framework for generating human behaviors based on object motions, utilizing conditional diffusion to generate hand joint positions as extra guidance. While this work lacks the integration of textual information, which can provide important semantic guidance. 
By contrast, our method fully leverages textual reasoning and incorporates hierarchical semantic and geometric features to enhance contextual motion generation.

\section{Method}
\begin{figure*}
 \vspace{-5ex}
    \centering
    \includegraphics[width=0.98\linewidth]{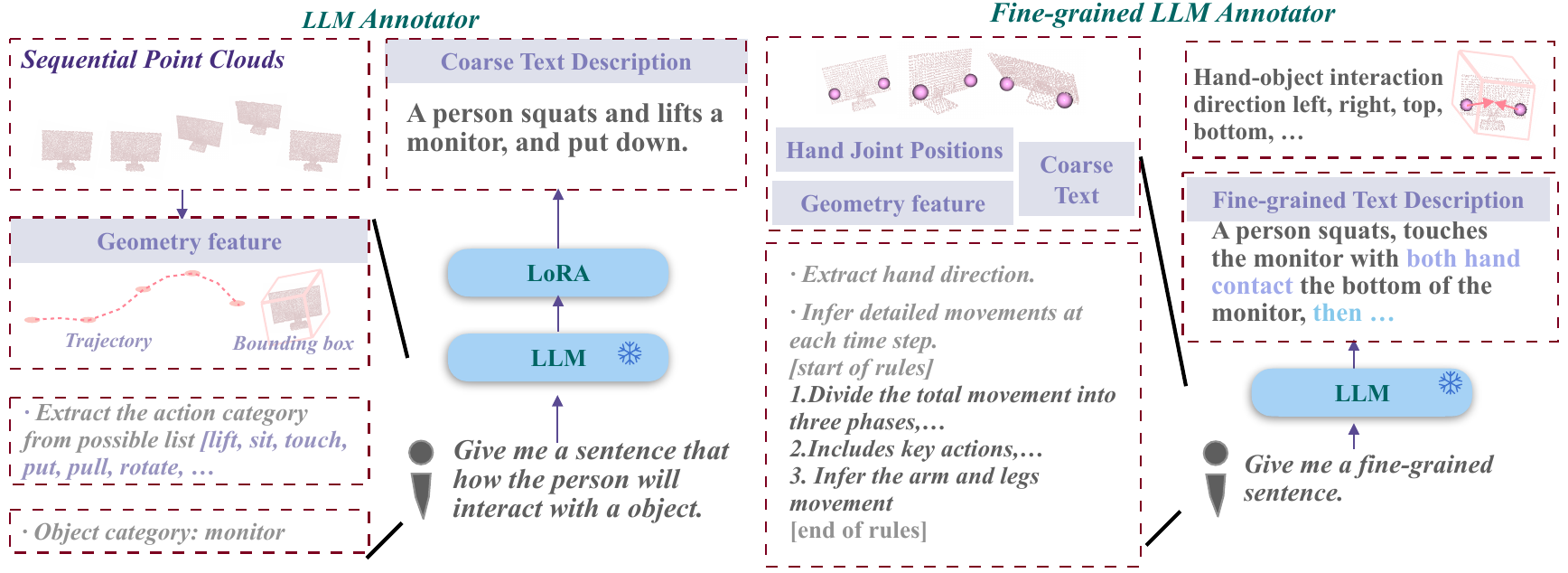}
    \vspace{-1ex}
    \caption{LLM Annotator pipeline. It first takes a sequential point cloud to infer a coarse text description. Then uses joint positions, the coarse text, and geometric features to generate a fine-grained sentence with a designed prompt.}
    \label{fig:LLM}
    \vspace{-4ex}
\end{figure*}

Our goal is to generate responsive human motions conditioned solely on the interactive target, represented as a 4D sequential point cloud. To achieve this, we introduce the LLM Annotator to provide semantic features, which are then processed in the following two stages: SemGeo Hierarchical Guidance Generation and SemGeo-guided Motion Generation. In the first stage, multi-level geometric information is generated in a coarse-to-fine manner. These generated guidance signals are then used in the second stage to guide the motion generation process, enhancing both semantic plausibility and geometric accuracy. The overall architecture is shown in Fig. \ref{fig:framework}.


\subsection{Data Representation}
\label{sec: 3.1}
\noindent\textbf{Motion Representations.}
We denote the human motion as $\mathbf{x}^h \in \mathbb{R} ^{L \times D}$, where $L$ represents the number of frames in the sequence and $D$ is the feature dimension, which includes the pelvis velocity, local joint positions, rotations and velocities of other joints in the pelvis space, as well as binary foot-ground contact labels, following the representation in MDM~\cite{mdm2022human}.

\noindent\textbf{Point Cloud Representations.}
To extract geometric features, point clouds are well-suited for capturing the dynamic changes of objects, providing a unified input modality for representing deformable or non-rigid interactive targets without the need for additional preprocessing.
The sequential point cloud is down-sampled to $N=1024$ points per frame, denoted as $\mathbf{P} \in \mathbb{R} ^{L \times N \times 3}$. We then adopt Basis Point Set~\cite{prokudin2019efficient} (BPS) representation, following the approach in \cite{li2023object}, to encode the object geometry. The process begins by uniform sampling the basis points from a unit ball with a radius of 1 meter, followed by calculating the minimum Euclidean distance from the basis points to their nearest neighbors. Finally, these distances are concatenated with the center position of the original point cloud. The resulting
point cloud representation for each frame is denoted as $\mathbf{p} \in \mathbb{R} ^{N \times 3 + 3}$. An MLP is then applied to project the BPS representation into a lower-dimensional space, yielding the sequential geometry features $\mathbf{F_{pc}} \in \mathbb{R} ^{L \times 256}$.

\noindent\textbf{Affordance Map Representations.}
The affordance map serves as an intermediate representation, providing crucial geometric information regarding which parts of the target are most likely to come into contact during the interaction. For each frame, we calculate the $\ell_2$ distance between each point and the skeleton joints, resulting in a per-frame distance map $d \in \mathbb{R}^{L \times N \times J}$, where $J$ is the number of skeleton joints. We then transform this distance field into a normalized distance map, Affordance, denoted as $\mathbf{A}$,
which encodes the spatial relationship between the target and the interactive human. The affordance map is computed as: 
\vspace{-1ex}
\begin{equation}
\vspace{-1ex}
\mathbf{Affordance}(n, j)=\exp \left(-\frac{1}{2} \frac{\mathbf{d}(n, j)}{\sigma^2}\right),
\end{equation}
where $\sigma$ is the normalizing factor.

\subsection{LLM Annotator}
Textual descriptions provide essential semantic information. Given the interactive targets, envisioning how to interact with them is crucial for generating realistic human motion. Previous work~\cite{mdm2022human,mdmprior,motiondiffuse} on text-to-motion has also validated the effectiveness of feature mapping from textual descriptions to motion generation.

To obtain textual guidance, we employ a Large Language Model (LLM) as an annotator in the initial stage to generate a coarse motion description. The whole pipeline is shown in Fig.~\ref{fig:LLM}. From the sequential point cloud, we extract the bounding box of the interactive target along with its movement trajectory. Utilizing a predefined list of actions and categories, we derive initial language-based guidance.
Specifically, we utilize the pre-trained LLaMA model \cite{touvron2023llama}, enhanced with LoRA \cite{hu2021lora} finetuning on given text-interaction pairs from \cite{li2023object}. 

More detailed, context-aware cues can capture the dynamic changes in contact, guiding the model to generate more precise body movements.  Leveraging the strong reasoning capabilities of LLMs, we design a coarse-to-fine automated language guidance annotation system. After generating initial textual guidance and predicting hand joint positions, we determine sequential contact information by calculating distances between predicted joint positions and interactive targets, inferring where the body should make contact with each part of the targets. The LLM then organizes this contact information into language descriptors (\eg, ``the left-hand contacts the lower left of the box''). Simultaneously, we let the LLM divide the interaction into three steps, with the textual descriptions reflecting changes in contact at each step, with also deducing finer body movements of the arms and legs. 

Thus, the generated textual descriptions assist in the motion generation task by providing semantic information. On the other hand, our entire pipeline also enables reasoning capabilities, allowing for a more comprehensive understanding and generation of human motion.

\subsection{SemGeo Hierarchical Guidance Generation}
The joint position $\mathbf{J_{h}} \in \mathcal{R}^{L \times J \times 3}$ provides precise spatial information, and the affordance map $\mathbf{A}$ offers coarse geometric clues. These two types of features are crucial for modeling interactions in a coarse-to-fine manner, thus we introduce a conditional diffusion model with dual-branch transformer to jointly generate contact information with capturing their mutual influence in the first stage.

We utilize CLIP\cite{clip} as text encoder to obtain the text feature $F_{text}$. At each step in the diffusion process, the model take the text feature $F_{text}$, point cloud feature $F_{pc} $, and noisy signals $x_t^J$ and $x_t^A$ as input and predict clean $x_0^J$ and $x_0^A$.
For JointTransformer, the inputs are concatenated together and then fed into the multi-head self-attention blocks followed with position-wise feedforward layer.
For AffordanceTransformer, inspired by \cite{wang2024move}, the affordance map is more closely related to the point cloud geometry. Therefore, we encode the point cloud feature $F_{pc}$ along with $x_t^A$, which act as the key and value in attention module. The concatenation of the language feature and diffusion step embeddings serves as the query. After passing through the cross-attention mechanism and multiple self-attention blocks, the refined point features are obtained.
To enhance the coarse to fine interactions and further refine the joint position, we introduce a mutual cross-attention mechanism. This take the output from Jointtransformer as query, output from AffordanceTransformer as key and value, which updates the hand joint position.

The conditional diffusion model learns the reverse diffusion process to generate clean data from a Gaussian noise $x_t$ over $T$ consecutive denoising steps.
Specifically, we use $c$ to represent the conditions, and the reverse diffusion process is modeled as:
\begin{equation}
p_\theta\left(x^{t-1} \mid x^t, \boldsymbol{c}\right):=\mathcal{N}\left(x^{t-1} ; \boldsymbol{\mu}_\theta\left(x^t, t, \boldsymbol{c}\right), \sigma_t^2 I\right).
\end{equation}
Finally, our model directly estimates the input signal. The training process optimizes from the reconstruction loss:
\begin{equation}
\mathcal{L}=\mathbb{E}_{\boldsymbol{x}^0, t}\left\|\hat{x}_\theta\left(x^t, t, \boldsymbol{c}\right)-x^0\right\|_1.
\end{equation}

\subsection{SemGeo-guided Motion Generation}
In the second stage, a denoising network architecture is used to generate full-body motions based on the predicted joint positions $\mathbf{J}_h'$, affordance map $\mathbf{A'}$, and text descriptions. 

\noindent\textbf{SemGeo Condition Module}
To effectively process and integrate these diverse features, we design SemGeo Condition Module to encode the combined features. 
After obtaining both coarse and fine-grained text descriptions, we use a text encoder to extract the semantic information. Since the language generated by the Fine-grained LLM Annotator contains phase-specific details and longer descriptions, we adopt LONGCLIP~\cite{zhang2025long}, which can better capture and represent fine-grained attributes without the length limitations inherent in CLIP. The extracted feature is $F'_{\text{text}}$.

To extract features from the affordance map, we first concatenate the point cloud feature $F_{pc}$ with the affordance map $\mathbf{A'}$, and then pass the combined input through a 3-layer MLP. The temporal transformer is applied to extract latent features $F$ over time, which helps to capture both spatial and temporal dependencies. The operation is formalized as:
\begin{equation}
    F = \text{TemporalTransformer}(MLP(F_{pc} \oplus \mathbf{A'}),
\end{equation}
where $\oplus$ denotes concatenation. 

We apply mutual cross-attention to extract the mutual features between joint positions and the affordance map. The joint feature, after being mapped to a higher dimensional space with MLP, serves as the query. The latent feature from the previous step is used as key and value in the attention mechanism. This interaction between joint positions and the affordance map enables the model to capture the relationship between joint movements and object geometry, improving motion prediction:
\vspace{-1ex}
\begin{equation}
    F_{fuison} = \text{CrossAttention}(MLP(\mathbf{J}_h')_q, F_k,F_v).
    \vspace{-1ex}
\end{equation}
The final condition $c$ is the concatenation result of $F_{text}$, $F'_{text}$, $F_{fuison}$.

This comprehensive feature representation ensures that the model has a rich, multi-dimensional understanding, capturing both coarse-to-fine semantic information from the text and spatial-temporal geometric information from the joint positions and affordances.

\noindent\textbf{Motion ControlNet}
Inspired by \cite{xie2023omnicontrol, wang2023intercontrol}, we introduce Motion ControlNet to generate high-fidelity motions conditioned on $c$. 
With MDM frozen during training, each transformer encoder layer in ControlNet~\cite{zhang2023adding} is linked to its MDM counterpart via a zero-initialized linear layer.

\noindent\textbf{Loss Guidance}
To refine our generated interactions, we employ joint guidance and foot guidance during sampling with classifier guidance.
The joint-based guidance loss aligns the generated global joint positions $\mathbf{J}_\text{pred}$ from the second stage with the target control joint positions $\mathbf{J'_h}$ from the first stage, ensuring consistency in the generated motion across both stages. 
Constraints are applied to the joints that are in contact with objects when the distance between the joint and the nearest point on the object is below a predefined threshold $\tau$. Specifically, the mask is defined as $\text{Mask} = \text{Dis}(\mathbf{J}_\text{h}, V) \leq \tau$.
The joint guidance function is then defined as:
\vspace{-1ex}
\begin{equation}
    L_{\text{joint}} = \frac{1}{J} \sum_{i=1}^L |\mathbf{J}_{\text{pred}_i} - \mathbf{J}'_{\text{h}_i}|_2 \cdot \text{Mask}_i.
\vspace{-2ex}
\end{equation}
The foot-stability loss \( L_{\text{foot}} \) 
is designed to ensure that the foot stays near the ground and penalize sudden changes in velocity to eliminate foot sliding:
\vspace{-1ex}
\begin{equation}
    L_{\text{foot}} = \frac{1}{L} \sum_{i=1}^{L} \left( (y_i - h_g)^2 + \alpha \mathbf{M}_c(v_i^2) + \beta \mathbf{M}_c(a_i^2) \right),
\vspace{-2ex}
\end{equation}
where $ y_i = \min(h_{l,i}, h_{r,i}) $ is the height of the lower foot for each frame and $h_g$ is the empirical values indicating contact with the ground.
 \( v_i = \| p_{i+1} - p_i \| \) is the foot velocity at frame \( i \), calculated from the foot position \( p \) and
\( a_i = \| v_{i+1} - v_i \| \) is the foot acceleration. $\mathbf{M}_c$ represents the mask for contact with the floor.  
\( \alpha \) and \( \beta \) are hyperparameters. 
Inspired by~\cite{wang2023intercontrol}, we employ L-BFGS for several iterations at each denoising step to update the posterior mean.
\begin{figure*}[ht]
    \centering
    \vspace{-1ex}
    \includegraphics[width=0.98\linewidth]{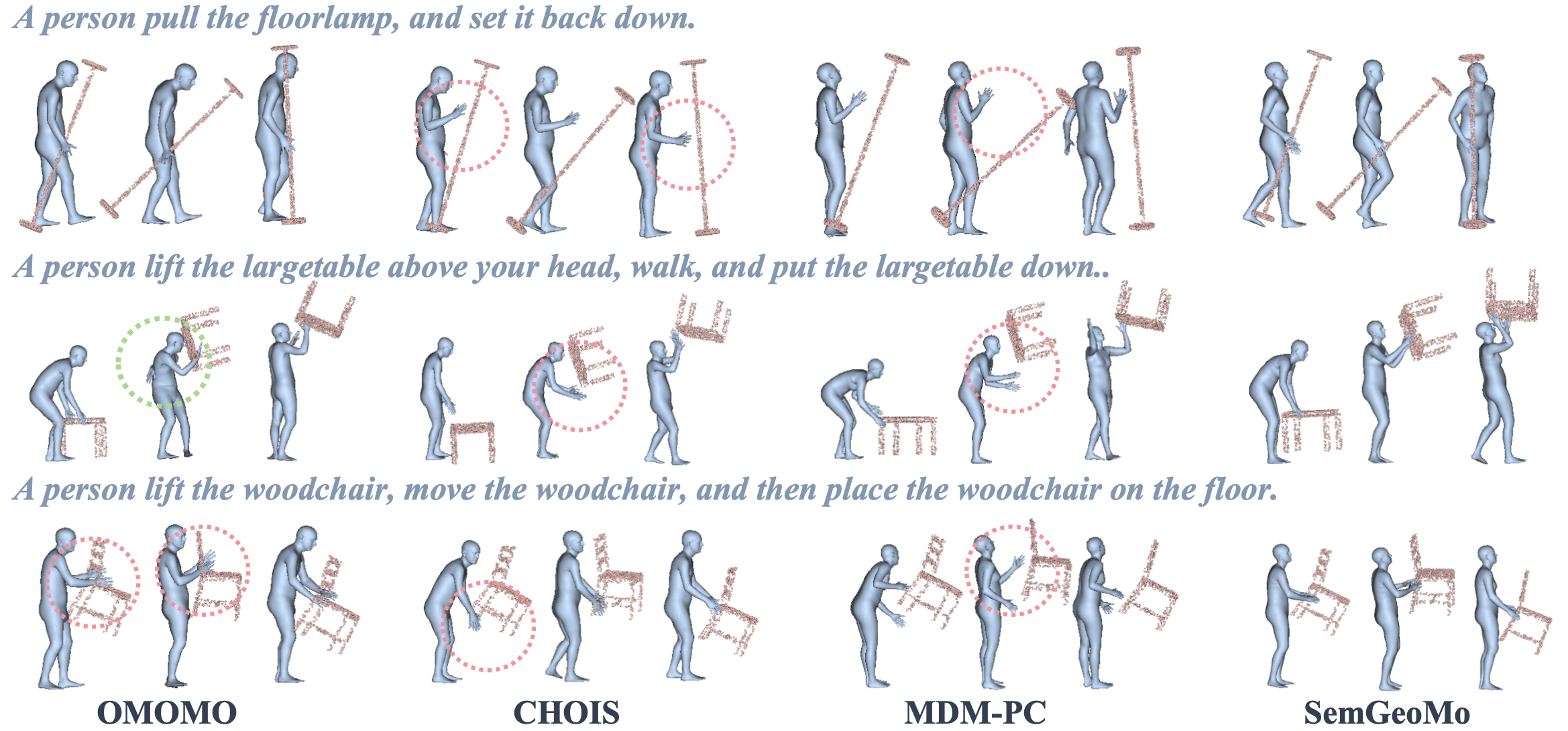}
    \caption{Qualitative results on the FullBodyManipulation dataset. We circle areas of low contact performance in pink and instances of contorted motion in green.}
    \label{fig:exp1}
    \vspace{-2ex}
\end{figure*}
\begin{table*}[h!]
\caption{Human motion generation result on FullBodyManipulation.}
\vspace{-1ex}
\resizebox{\linewidth}{!}{ 
\begin{tabular}{c|l|cccccccccccc}
\rowcolor[HTML]{D9E1F4} 
\hline
\multicolumn{1}{l|}{} &               & HandJPE$\downarrow$                                & MPJPE   $\downarrow$                               & $C_{prec} \uparrow$                       & $C_{rec}\uparrow$                       & $C_{acc}\uparrow$ & c\%$\uparrow$    & F1 $\uparrow$   & FID $\downarrow$ & R-score $\uparrow$   & Diversity $\uparrow$ & FS $\downarrow$  \\ \hline
& SceneDiff ~\cite{scenediff}    &   95.38                                        &    19.84                                    &     0.64                             &             0.19                     &  0.45            &  0.18   & 0.27       & 1.64    &    0.59  & 9.86   &   0.38   \\ \cline{2-13} 
& OMOMO ~\cite{li2023object}      & 33.18                                   & 18.06                                  & 0.77                             & 0.71                           & 0.74           & 0.61  &0.75     & 1.98   & 0.38      & 8.99& 0.50   \\ \hline
\multirow{-4}{*}{w/o text}                   
& MDM-PC~\cite{mdm2022human}     &     51.35	&18.25&	0.71	&0.41	&0.62    &    0.33 &   0.49    &    \textbf{0.65} &     0.57     & 9.52 &   0.51  \\ \cline{2-13} 
 & CHOIS  \cite{li2023controllable}       &   31.68                                        &  17.12                                      & 0.76                              & 0.58                             & 0.61         & 0.52  &    0.59 &  2.27 &0.49 & 6.04& 0.47     \\ \cline{2-13} 
& AffordMotion \cite{wang2024move} &             98.66                              &    25.34                                    &         0.45                         &     0.14                             &      0.31        &   0.13  &   0.16    &  4.71   &  0.45     & 8.15  & \textbf{0.43}     \\ \cline{2-13} 
\multirow{-3}{*}{w GT text}  & SemGeoMo       & \textbf{27.84}                           & \textbf{16.62}                               & \textbf{0.84}                         & \textbf{0.74}                            & \textbf{0.85}       & \textbf{0.66}    & \textbf{0.77} &   1.17  &    \textbf{0.66}     &\textbf{10.15}& 0.57     \\\hline
\multirow{-1}{*}{w Gen text}  & SemGeoMo          &   30.35                                &        17.98                    &      0.82              &         0.74                 & 0.82    & 0.66    & 0.74 &    1.05 & 0.64    & 9.78   &   0.47   \\ \hline
\end{tabular}}
\label{tab:omomo1}
\vspace{-2ex}
\end{table*}
\section{Experiment}
\subsection{Datasets}

\noindent\textbf{FullBodyManipulation \cite{li2023object}}
takes a total duration of approximately
10 hours. It provides, paired object and human motion, including interactions of 17 subjects with 15 different objects with text descriptions. We follow the official train/test split for evaluation.

\noindent\textbf{BEHAVE~\cite{bhatnagar2022behave}} consists of the interactions of 8 subjects with 20 different objects. We follow the official train/test split provided by BEHAVE. \cite{peng2023hoi} provides the text annotations while we observe that these annotations merely combine action labels and categories of motion, offering phase-specific details. We provide a revised version of the annotations with phased interactions that more clearly describe the motions.

\noindent\textbf{IMHD}$^2$~\cite{Zhao_2024_CVPR} and \textbf{HoDome}~\cite{zhang2023neuraldome} are challenging 3D human-object interaction datasets for motion capture, covering interactions between 10 objects and 10 subjects, and 23 diverse objects with 10 subjects, respectively. We annotate the text descriptions using our LLM annotation pipeline to facilitate further text-driven human-object interaction studies and enable comparisons with other text-guided methods.

\subsection{Evaluation Metrics}
We mainly follow the metrics in OMOMO~\cite{li2023object} and MDM~\cite{mdm2022human}.
\textbf{HandJPE} and \textbf{MPJPE} represent mean hand joint position errors, and mean per-joint position errors, computed using the Euclidean distance between the predicted and ground truth in centimeters (cm).
For measuring the interaction quality, we employ contact metrics including precision (\textbf{$\mathbf{C}_{prec}$}), recall (\textbf{$\mathbf{C}_{recall}$}), accuracy ({$\mathbf{C}_{acc}$}) and \textbf{F1 score} following \cite{li2023object}. The contact percentage ($c\%$) is the proportion of frames where contact is detected. \textbf{FID} measures the distance of the generated motion distribution to the ground truth distribution in latent space. \textbf{R-score} measures the text and motion matching accuracy and \textbf{Diversity} measures the generation diversity.
\textbf{FS} represents foot sliding metric and is computed following \cite{he2022nemf}.

\begin{table*}[h!]
\caption{Human motion generation result on Behave.}
\vspace{-1ex}
\resizebox{\linewidth}{!}{ 
\begin{tabular}{c|l|ccccccccccc}
\rowcolor[HTML]{D9E1F4} 
\hline
\multicolumn{1}{l|}{} &               & HandJPE$\downarrow$                                & MPJPE   $\downarrow$                               & $C_{prec} \uparrow$                       & $C_{rec}\uparrow$                       & $C_{acc}\uparrow$ & c\%$\uparrow$    & F1 $\uparrow$   & FID $\downarrow$ & R-score $\uparrow$   & Diversity $\uparrow$ & FS $\downarrow$  \\ \hline
& SceneDiff~\cite{scenediff}    &  51.58                                        &   18.25                                &     0.73                   &             0.38                   &  0.40           &  0.32   & 0.47      &   1.69  &    0.13 & 5.32   &   0.33  \\ \cline{2-13} 
& OMOMO~\cite{li2023object}      &    45.35                               & 21.56                                  &       0.71                       &      0.60                     &   0.61        & 0.60 &  0.62 & 1.94   &  0.14    &5.11 &0.42   \\ \hline
\multirow{-4}{*}{w/o text}                   
& MDM-PC~\cite{mdm2022human}     &     35.41                           &         18.61                         &      0.73                       &   0.48                             &       0.51    &    0.47 &   0.57   &  1.52  &   0.10      & 5.45 &   \textbf{0.32}  \\ \cline{2-13} 
 & CHOIS~\cite{li2023controllable}       &  36.75   & 18.17  &                  0.72                     & 0.41                                     &  0.43                            &   0.41                          &   0.51       &    2.26 &  0.13   &  5.02&  0.46 \\ \cline{2-13} 
& AffordMotion~\cite{wang2024move} &     55.65                                      &    19.16                                    &   0.72                               &  0.23                                & 0.28             &  0.25   & 0.32    &   1.92  &  0.13    & 4.38    & 0.51       \\ \cline{2-13} 
\multirow{-3}{*}{w Gen text}  & SemGeoMo        &  \textbf{27.91}                        & \textbf{16.22}                              & \textbf{0.84}                      & \textbf{0.67}                            & \textbf{0.67}       & \textbf{0.66}    & \textbf{0.74} &  \textbf{1.47}   &   \textbf{0.15}       & \textbf{5.64} & 0.52     \\\hline
\end{tabular}}
\label{tab:behave}
\vspace{-1ex}
\end{table*}

\begin{table*}[h!]
\caption{Human motion generation result on IMHD$^2$.}
\vspace{-2ex}
\resizebox{\linewidth}{!}{ 
\begin{tabular}{c|l|cccccccccccc}
\rowcolor[HTML]{D9E1F4} 
\hline
\multicolumn{1}{l|}{} &                & HandJPE$\downarrow$                                & MPJPE   $\downarrow$                               & $C_{prec} \uparrow$                       & $C_{rec}\uparrow$                       & $C_{acc}\uparrow$ & c\%$\uparrow$    & F1 $\uparrow$   & FID $\downarrow$ & R-score $\uparrow$  & Diversity $\uparrow$ & FS $\downarrow$  \\ \hline
& SceneDiff~\cite{scenediff}     &  82.01                                        &   25.08                                &     0.45                    &             0.18                    &  0.21           &  0.16   & 0.22      &  1.89   &  0.15     &   5.22 &   0.57  \\ \cline{2-13} 
& OMOMO~\cite{li2023object}      & 39.40                                   & 23.36                                  & 0.58                             & 0.39                           & 0.43           & 0.41 &0.42    &2.09&  \textbf{0.16}    & 4.56 & 0.55  \\ \hline
\multirow{-4}{*}{w/o text}                   
& MDM-PC~\cite{mdm2022human}     &     63.57&23.81	&0.48&	0.24	                           &       0.28     &    0.30 &   0.24    &  1.73   &  0.14     &   5.23 & 0.59  \\ \cline{2-13} 
 & CHOIS  \cite{li2023controllable}       &   44.92                                       & 25.09                                       &  0.56                             &0.31                              &  0.35        & 0.31 &   0.32  &  2.67 &   0.11     & 4.45 &   0.53   \\ \cline{2-13} 
& AffordMotion \cite{wang2024move} &     75.32                                      &    24.37                                    &   0.55                               &    0.16                              &    0.22          &   0.16 &   0.21    &   2.88  & 0.12        & 4.48  &  \textbf{0.42}    \\ \cline{2-13} 
\multirow{-3}{*}{w Gen text}  & SemGeoMo        &  \textbf{35.43}                         & \textbf{20.85}                               & \textbf{0.72}                       & \textbf{0.49}                            & \textbf{0.51}       & \textbf{0.49}    & \textbf{0.52} &  \textbf{1.64}   & 0.14      & \textbf{5.35}& 0.49     \\\hline

\end{tabular}}
\label{tab:imhd}
\vspace{-1ex}
\end{table*}

\begin{table*}[h!]
\caption{Human motion generation result on HoDome.}
\vspace{-1ex}
\resizebox{\linewidth}{!}{ 
\begin{tabular}{c|l|ccccccccccc}
\rowcolor[HTML]{D9E1F4} 
\hline
\multicolumn{1}{l|}{} &               & HandJPE$\downarrow$                                & MPJPE   $\downarrow$                               & $C_{prec} \uparrow$                       & $C_{rec}\uparrow$                       & $C_{acc}\uparrow$ & c\%$\uparrow$    & F1 $\uparrow$   & FID $\downarrow$ & R-score $\uparrow$   & Diversity $\uparrow$ & FS $\downarrow$  \\ \hline
& SceneDiff~\cite{scenediff}  & 107.50       &   29.53 &   0.21    &   0.11    &      0.16       & 0.13   & 0.14  &   4.88 & 0.08& 4.72     &   0.60     \\ \cline{2-13} 
& OMOMO~\cite{li2023object}   & 86.12      &   27.07 &   0.42    &   0.23   &      0.31       & 0.22   & 0.25  &5.47    &0.10 &   4.93   &   0.31      \\ \hline
\multirow{-4}{*}{w/o text}                   
& MDM-PC~\cite{mdm2022human} & 95.91       &   26.29 &   0.25   &   0.13    &      0.18       & 0.14   & 0.15 &   \textbf{3.64} & 0.13 & 4.89    &   0.67     \\ \cline{2-13} 
 & CHOIS~\cite{li2023controllable}    & 76.74      &   \textbf{24.17} &   0.55    &   0.12   &      0.26       & 0.11  & 0.18  & 6.12   &0.11 &  4.14    &   \textbf{0.28}  \\ \cline{2-13} 
& AffordMotion~\cite{wang2024move}& 94.24       &   29.31  &   0.49     &   0.11     &      0.15        & 0.10   & 0.13 &5.29 &0.11 &5.14 & 0.45   \\ \cline{2-13} 
\multirow{-3}{*}{w Gen text}  & SemGeoMo     &  \textbf{44.22}       &   24.28  &   \textbf{0.78}     &   \textbf{0.47}     &      \textbf{0.47}        & \textbf{0.45}   & \textbf{0.54}  &  4.29  & \textbf{0.13} &   \textbf{5.22}   &   0.35    \\\hline
\end{tabular}}
\vspace{-2ex}
\label{tab:dome}
\end{table*}
\subsection{Results}
\noindent\textbf{Baselines}
OMOMO~\cite{li2023object} is the only work which is aligned with our setting. In addition, we adapt several related works, including SceneDiff~\cite{scenediff}, MDM-PC~\cite{mdm2022human}, AffordMotion~\cite{wang2024move}, and CHOIS \cite{li2023controllable}, to fit our problem setting.  SceneDiff~\cite{scenediff} utilizes a diffusion model conditioned on static scenes, so we modified the conditional model to accommodate sequential point cloud inputs.
We modified the text-to-motion generation work MDM as MDM-PC, incorporating our sequential point cloud representation. CHOIS and AffordMotion are conditioned on both textual descriptions and scenarios. CHOIS requires a sequence of object states in 2D waypoints, which we modify this part into 3D states. AffordMotion predicts the affordance map and subsequently generates motion in scenes, we adopted this approach for our interactive target setting. Note that for the aforementioned works, since FullBodyManipulation provide the textual description, we provided ground-truth (GT) text as input, and compared the variant of our model under the same conditions with ground-truth text as well. 

\noindent\textbf{Results on the FullBodyManipulation Dataset.}
The results on the FullBodyManipulation dataset are illustrated in Tab.~\ref{tab:omomo1}, and we provide a visualization comparison in Fig. \ref{fig:exp1}. 
SceneDiff, MDM-PC, and CHOIS directly use the original point cloud as a condition in a single stage, the dynamic nature of the point cloud makes it challenging to infer low-level contact information, resulting in a significant drop in contact metrics. Even though MDM incorporates text guidance and improves performance on FID, the contact accuracy remains poor. CHOIS applies a human-object loss to refine contact, but such improvement in performance is limited. OMOMO predicts joint positions as the first-stage output, but without textual guidance, it performs poorly on FID and R-score, struggling to capture realistic interaction motions. Distortions and abnormal movements may occur, as shown in Fig. \ref{fig:exp1} circled in green. AffordMotion uses the affordance map for geometric guidance, while this approach lacks sufficient detail for fine-grained human-object interactions. In contrast, we leverage both semantic and geometric information by generating textual descriptions and predicting coarse-to-fine contact information, resulting in superior performance. Compared to the variant, which utilizes ground-truth text information as a condition, our model achieves comparable performance, demonstrating the efficiency and accuracy of the language description generated by our method.

\noindent\textbf{Results on the Behave and IMHD$^2$ Dataset.}
We further conduct experiments on the BEHAVE and IMHD$^2$ datasets to validate our model’s performance in Tab.~\ref{tab:behave} and Tab.~\ref{tab:imhd}. Notably, the IMHD$^2$ dataset includes more challenging interactions, such as sports activities, which make precise contact and accurate motion generation more difficult. However, our proposed method still outperforms other approaches in these scenarios.

\begin{figure}
    \centering
\includegraphics[width=1.0\linewidth]{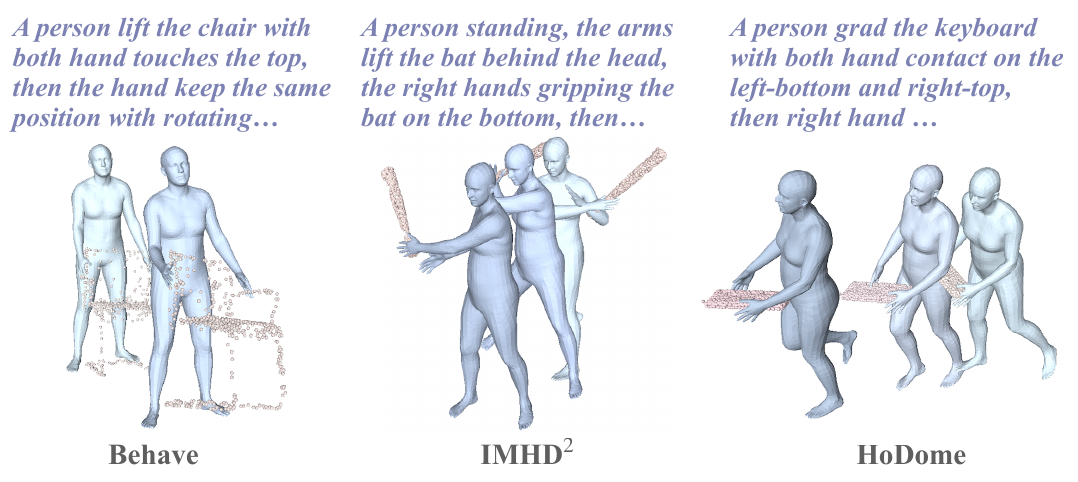}
 \vspace{-4ex}
    \caption{Visulization on more datasets.}
    \label{fig:vis-other}
    \vspace{-2ex}
\end{figure}

\noindent\textbf{Results on the HoDome Dataset.}
To test our model's ability to generate new scenarios, we further conduct the experiment on totally unseen objects with directly sampling on HoDome~\cite{zhang2023neuraldome}. The result is illustrated in Tab.~\ref{tab:dome}, our proposed method outperforms other methods. The visualization results are illustrated in Fig.~\ref{fig:vis-other}.

\begin{figure}
    \centering
\includegraphics[width=1.0\linewidth]{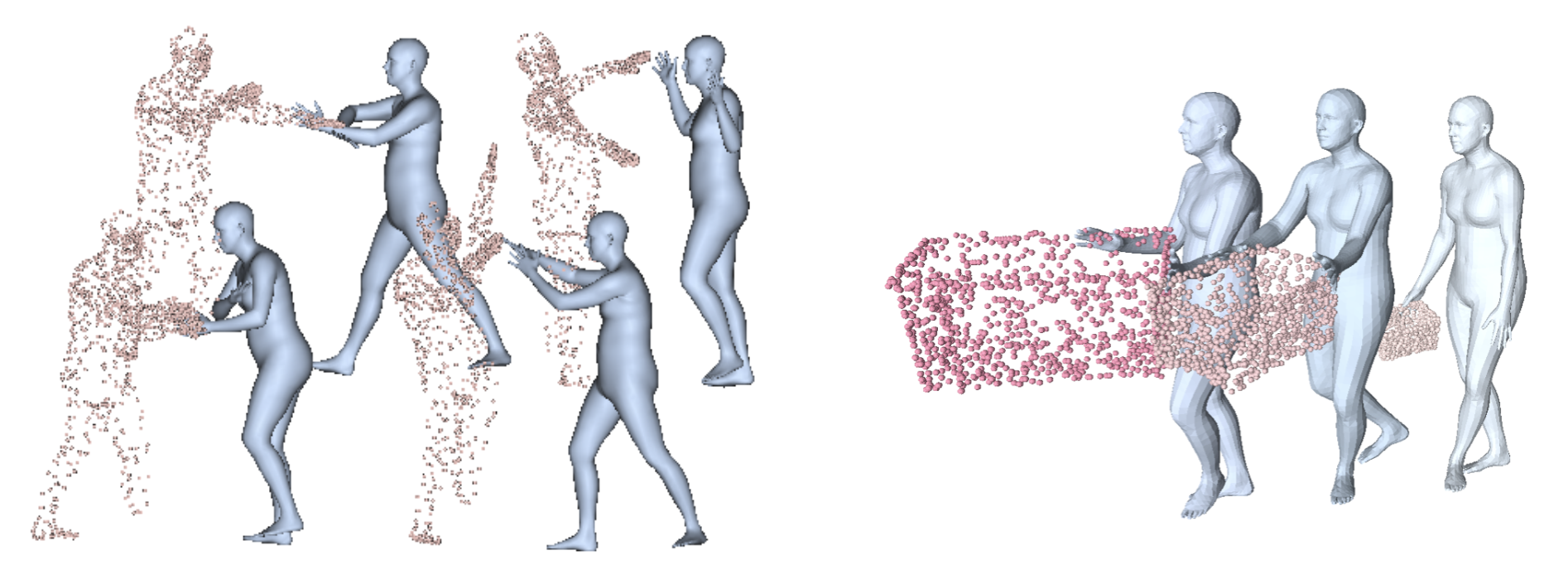}
    \caption{Extension on other scenarios. (a) is the interaction with humans and (b) is the interaction with objects of varying sizes. }
    \label{fig:extension}
     \vspace{-2ex}
\end{figure}

\begin{table*}[h!]
\caption{Ablation studies on the impact of semantic and geometric information with our model design.}
\vspace{-1ex}
\resizebox{\linewidth}{!}{ 
\begin{tabular}{c|cccccccccccc}
\rowcolor[HTML]{D9E1F4} 
\hline
 & HandJPE$\downarrow$                                & MPJPE   $\downarrow$                               & $C_{prec} \uparrow$                       & $C_{rec}\uparrow$                       & $C_{acc}\uparrow$ & c\%$\uparrow$    & F1 $\uparrow$   & FID $\downarrow$ & R-score $\uparrow$   & Diversity $\uparrow$ & FS $\downarrow$  \\ \hline\hline
w/o affordanc map                        &  29.89          &    18.85  &   0.80      &   0.71     &      0.81        & 0.63    & 0.73   &   2.21  & 0.62&9.84        &   0.58 \\ \hline
w/o joint                         &            31.23    &    20.60      &    0.76     &  0.70          &       0.78  & 0.61    &     0.72   & 3.52&0.51&8.15&  0.74 \\ \hline
w/o attention                       & 29.18         &   19.47            &       0.81     &         0.66    &   0.78  &   0.59 &    0.70 &     9.27&0.44&6.84&0.70  \\ \hline\hline
 w/o text     &  30.36                           & 24.44                               & 0.78                         & 0.62                            & 0.72       & 0.54    
 & 0.64 &   1.78  &  0.41     & 9.40 
 & 0.61 \\ \hline
 w/o fine-grained text     &  27.84                           & \textbf{16.62}                               & 0.84                         & 0.74                            & 0.85       & 0.66    & 0.77 &   1.17  &    0.66    &\textbf{10.15} & 0.57 \\ \hline\hline
Full     &  \textbf{27.97}                           & 17.01                               & \textbf{0.84}                         & \textbf{0.75}                            & \textbf{0.86}       & \textbf{0.67}    & \textbf{0.77} & \textbf{1.03}   & \textbf{0.68} & 10.05 & \textbf{0.46}
 \\ \hline
\textcolor{gray!70}{Full-GT}         &    \textcolor{gray!70}{6.44  }                              &        \textcolor{gray!70}{ 13.61  }                    &     \textcolor{gray!70}{  0.88 }                  &         \textcolor{gray!60}{ 0.82 }                   & \textcolor{gray!70}{ 0.93}   & \textcolor{gray!70}{0.70} &  \textcolor{gray!70}{0.84 }  &  \textcolor{gray!70}{0.97} &  \textcolor{gray!70}{ 0.70}  &  \textcolor{gray!70}{10.48} &    \textcolor{gray!70}{0.58}   \\  \hline
\end{tabular}}
\label{tab:model}
\end{table*}

\begin{table*}[h!]
\caption{Ablation study on guidance loss.}
\vspace{-2ex}
\resizebox{\linewidth}{!}{ 
\begin{tabular}{c|c|cccccccccccc}
\rowcolor[HTML]{D9E1F4} 
\hline
$L_{Joint}$ & $L_{FS}$   & HandJPE$\downarrow$                                & MPJPE   $\downarrow$                               & $C_{prec} \uparrow$                       & $C_{rec}\uparrow$                       & $C_{acc}\uparrow$ & c\%$\uparrow$    & F1 $\uparrow$   & FID $\downarrow$ & R-score $\uparrow$  & Diversity $\uparrow$ & FS $\downarrow$  \\ \hline
{\Checkmark}          &   & 27.84       &        16.97    &       0.83    &     0.75       &   0.86           & 0.67    & 0.75   &  1.15   &   0.65   & 10.13  &0.61   \\\hline
           & {\Checkmark}         &      85.28      &   29.75    & 0.62           &    0.25        &    0.35    &  0.24     &  0.22   &   0.93 &  0.66    &10.05 & 0.31  \\\hline
{\Checkmark}   &   {\Checkmark}   &  27.84                           & 16.62                               & 0.84                         & 0.74                            & 0.85       & 0.66    & 0.77 &   1.17  &    0.66      &10.15& 0.57    \\\hline   
\end{tabular}
}
\label{tab:loss}
\vspace{-2ex}
\end{table*}

\begin{table}[h!]
\caption{Text generation result on FullBodyManipulation.}
\vspace{-1ex}
\centering
\begin{tabular}{cccc}
\hline
\textbf{} BLEU-4 & ROUGE
-1 & ROUGE-2 & ROUGE-L \\ \hline
87.21  & 90.91   & 83.79   & 89.39   \\ \hline
\end{tabular}
\label{tab:text}
\vspace{-2ex}
\end{table}

\subsection{Ablation Studies}
\noindent\textbf{Ablation studies on the impact of semantic and geometric information.}
To verify the effectiveness of our multi-level guidance, we conduct ablation studies to assess the impact of each component in Tab.~\ref{tab:model}.
The multi-level geometric guidance and semantic information enhance human motion generation, reflected in the improvement of the contact and FID\&R-score indicators, respectively.
We also compare our designed SemGeo conditional module with cross-attention. The results show that cross-attention significantly improves performance by effectively sharing information between coarse-to-fine features.
During implementation, we predict the hand joint and compare with generation process with ground truth position (denoted as Full-GT), it is worth noting that when the hand is fixed to the exact ground truth, this may reduce the FS score due to some foot sliding in the generated motion. Our results are comparable to the Full-GT setting, indicating that our joint and affordance predictions are sufficiently accurate to provide effective guidance for generating plausible results.

\noindent\textbf{Analysis on text annotation.} To verify the correctness of our generated text, we calculate BLEU~\cite{papinesi2002bleu} and ROUGE~\cite{lin2005recall} scores, compared with the textual information provided by FullBodyManipulation in Tab.~\ref{tab:text}. These metrics illustrate the accuracy of our reasoning results. As shown in Tab.~\ref{tab:omomo1}, we achieve performance comparable to using ground truth text, further demonstrating the effectiveness of our textual descriptions. Additionally, the results on both coarse and fine-grained text highlight the effectiveness of incorporating multi-level semantic information. 

\noindent\textbf{Ablation studies on loss guidance.}
We conduct experiments on the loss guidance described in Tab.~\ref{tab:loss}. The joint loss plays a key role in improving contact performance, while the foot loss enhances the feasibility of the generated results. These improvements are reflected in the enhancement of FID and FS metrics.

\subsection{Extension}
We further evaluate our model on more challenging scenarios, including complex human-human interactions and shape-varying object manipulations, as Fig.~\ref{fig:extension} shows. 
The varying size of objects are simulated to assess how well our model generates responsive and adaptive interactions. The results demonstrate the feasibility of our approach, with generating realistic human responsive motions.

\section{Conclusion}
In this work, we introduce SemGeoMo, a novel method for generating responsive human motions and corresponding textual descriptions based on dynamic interactive targets.
We design an automated text annotator to provide semantic information. By integrating text-affordance-joint semantic and geometric guidance, SemGeoMo ensures the semantic coherence of the generated text and the geometric precision of the corresponding motion. Our method achieves state-of-the-art performance on three benchmarks and demonstrates generalization abilities on an unseen dataset.

\newpage
{
    \small
    \bibliographystyle{ieeenat_fullname}
    \bibliography{main}

\begin{thebibliography}{51}
\providecommand{\natexlab}[1]{#1}
\providecommand{\url}[1]{\texttt{#1}}
\expandafter\ifx\csname urlstyle\endcsname\relax
  \providecommand{\doi}[1]{doi: #1}\else
  \providecommand{\doi}{doi: \begingroup \urlstyle{rm}\Url}\fi

\bibitem[Ara{\'u}jo et~al.(2023)Ara{\'u}jo, Li, Vetrivel, Agarwal, Wu, Gopinath, Clegg, and Liu]{araujo2023circle}
Joao~Pedro Ara{\'u}jo, Jiaman Li, Karthik Vetrivel, Rishi Agarwal, Jiajun Wu, Deepak Gopinath, Alexander~William Clegg, and Karen Liu.
\newblock Circle: Capture in rich contextual environments.
\newblock In \emph{Proceedings of the IEEE/CVF Conference on Computer Vision and Pattern Recognition}, pages 21211--21221, 2023.

\bibitem[Bhatnagar et~al.(2022)Bhatnagar, Xie, Petrov, Sminchisescu, Theobalt, and Pons-Moll]{bhatnagar2022behave}
Bharat~Lal Bhatnagar, Xianghui Xie, Ilya~A Petrov, Cristian Sminchisescu, Christian Theobalt, and Gerard Pons-Moll.
\newblock Behave: Dataset and method for tracking human object interactions.
\newblock In \emph{Proceedings of the IEEE/CVF Conference on Computer Vision and Pattern Recognition}, pages 15935--15946, 2022.

\bibitem[Cen et~al.(2024)Cen, Pi, Peng, Shen, Yang, Zhu, Bao, and Zhou]{cen2024generating}
Zhi Cen, Huaijin Pi, Sida Peng, Zehong Shen, Minghui Yang, Shuai Zhu, Hujun Bao, and Xiaowei Zhou.
\newblock Generating human motion in 3d scenes from text descriptions.
\newblock In \emph{Proceedings of the IEEE/CVF Conference on CVPR}, pages 1855--1866, 2024.

\bibitem[Chen et~al.(2023)Chen, Jiang, Liu, Huang, Fu, Chen, and Yu]{mld}
Xin Chen, Biao Jiang, Wen Liu, Zilong Huang, Bin Fu, Tao Chen, and Gang Yu.
\newblock Executing your commands via motion diffusion in latent space.
\newblock In \emph{Proceedings of the IEEE/CVF CVPR}, pages 18000--18010, 2023.

\bibitem[Cong et~al.(2024)Cong, Wang, Dou, Ren, Yin, Cheng, Sun, Long, Zhu, and Ma]{cong2024laserhuman}
Peishan Cong, Ziyi Wang, Zhiyang Dou, Yiming Ren, Wei Yin, Kai Cheng, Yujing Sun, Xiaoxiao Long, Xinge Zhu, and Yuexin Ma.
\newblock Laserhuman: Language-guided scene-aware human motion generation in free environment.
\newblock \emph{arXiv preprint arXiv:2403.13307}, 2024.

\bibitem[Dabral et~al.(2023{\natexlab{a}})Dabral, Mughal, Golyanik, and Theobalt]{Mofusion}
Rishabh Dabral, Muhammad~Hamza Mughal, Vladislav Golyanik, and Christian Theobalt.
\newblock Mofusion: A framework for denoising-diffusion-based motion synthesis.
\newblock In \emph{Proceedings of the IEEE/CVF Conference on Computer Vision and Pattern Recognition}, pages 9760--9770, 2023{\natexlab{a}}.

\bibitem[Dabral et~al.(2023{\natexlab{b}})Dabral, Mughal, Golyanik, and Theobalt]{dabral2023mofusion}
Rishabh Dabral, Muhammad~Hamza Mughal, Vladislav Golyanik, and Christian Theobalt.
\newblock Mofusion: A framework for denoising-diffusion-based motion synthesis.
\newblock In \emph{Proceedings of the IEEE/CVF conference on computer vision and pattern recognition}, pages 9760--9770, 2023{\natexlab{b}}.

\bibitem[Diller and Dai(2024)]{diller2024cg}
Christian Diller and Angela Dai.
\newblock Cg-hoi: Contact-guided 3d human-object interaction generation.
\newblock In \emph{Proceedings of the IEEE/CVF Conference on Computer Vision and Pattern Recognition}, pages 19888--19901, 2024.

\bibitem[Hassan et~al.(2019)Hassan, Choutas, Tzionas, and Black]{hassan2019resolving}
Mohamed Hassan, Vasileios Choutas, Dimitrios Tzionas, and Michael~J Black.
\newblock Resolving 3d human pose ambiguities with 3d scene constraints.
\newblock In \emph{Proceedings of the IEEE/CVF international conference on computer vision}, pages 2282--2292, 2019.

\bibitem[He et~al.(2022)He, Saito, Zachary, Rushmeier, and Zhou]{he2022nemf}
Chengan He, Jun Saito, James Zachary, Holly Rushmeier, and Yi Zhou.
\newblock Nemf: Neural motion fields for kinematic animation.
\newblock \emph{Advances in Neural Information Processing Systems}, 35:\penalty0 4244--4256, 2022.

\bibitem[Ho and Salimans(2022)]{ho2022classifier}
Jonathan Ho and Tim Salimans.
\newblock Classifier-free diffusion guidance.
\newblock \emph{arXiv preprint arXiv:2207.12598}, 2022.

\bibitem[Ho et~al.(2020)Ho, Jain, and Abbeel]{ho2020denoising}
Jonathan Ho, Ajay Jain, and Pieter Abbeel.
\newblock Denoising diffusion probabilistic models.
\newblock \emph{Advances in neural information processing systems}, 33:\penalty0 6840--6851, 2020.

\bibitem[Hu et~al.(2021)Hu, Shen, Wallis, Allen-Zhu, Li, Wang, Wang, and Chen]{hu2021lora}
Edward~J Hu, Yelong Shen, Phillip Wallis, Zeyuan Allen-Zhu, Yuanzhi Li, Shean Wang, Lu Wang, and Weizhu Chen.
\newblock Lora: Low-rank adaptation of large language models.
\newblock \emph{arXiv preprint arXiv:2106.09685}, 2021.

\bibitem[Huang et~al.(2023)Huang, Wang, Li, Jia, Liu, Zhu, Liang, and Zhu]{scenediff}
Siyuan Huang, Zan Wang, Puhao Li, Baoxiong Jia, Tengyu Liu, Yixin Zhu, Wei Liang, and Song-Chun Zhu.
\newblock Diffusion-based generation, optimization, and planning in 3d scenes.
\newblock In \emph{Proceedings of the IEEE/CVF Conference on Computer Vision and Pattern Recognition}, pages 16750--16761, 2023.

\bibitem[Jiang et~al.(2023{\natexlab{a}})Jiang, Chen, Liu, Yu, Yu, and Chen]{jiang2023motiongpt}
Biao Jiang, Xin Chen, Wen Liu, Jingyi Yu, Gang Yu, and Tao Chen.
\newblock Motiongpt: Human motion as a foreign language.
\newblock \emph{Advances in Neural Information Processing Systems}, 36:\penalty0 20067--20079, 2023{\natexlab{a}}.

\bibitem[Jiang et~al.(2024{\natexlab{a}})Jiang, Chen, Liu, Yu, Y, and Chen]{jiang2024motiongpt}
Biao Jiang, Xin Chen, Wen Liu, Jingyi Yu, Gang Y, and Tao Chen.
\newblock Motiongpt: Human motion as a foreign language.
\newblock \emph{Advances in Neural IPS}, 36, 2024{\natexlab{a}}.

\bibitem[Jiang et~al.(2022)Jiang, Liu, Cao, Cui, Chen, Wang, Zhu, and Huang]{jiang2022chairs}
Nan Jiang, Tengyu Liu, Zhexuan Cao, Jieming Cui, Yixin Chen, He Wang, Yixin Zhu, and Siyuan Huang.
\newblock Chairs: Towards full-body articulated human-object interaction.
\newblock \emph{arXiv preprint arXiv:2212.10621}, 3, 2022.

\bibitem[Jiang et~al.(2023{\natexlab{b}})Jiang, Liu, Cao, Cui, Zhang, Chen, Wang, Zhu, and Huang]{jiang2023full}
Nan Jiang, Tengyu Liu, Zhexuan Cao, Jieming Cui, Zhiyuan Zhang, Yixin Chen, He Wang, Yixin Zhu, and Siyuan Huang.
\newblock Full-body articulated human-object interaction.
\newblock In \emph{Proceedings of the IEEE/CVF International Conference on Computer Vision}, pages 9365--9376, 2023{\natexlab{b}}.

\bibitem[Jiang et~al.(2024{\natexlab{b}})Jiang, Zhang, Li, Ma, Wang, Chen, Liu, Zhu, and Huang]{jiang2024scaling}
Nan Jiang, Zhiyuan Zhang, Hongjie Li, Xiaoxuan Ma, Zan Wang, Yixin Chen, Tengyu Liu, Yixin Zhu, and Siyuan Huang.
\newblock Scaling up dynamic human-scene interaction modeling.
\newblock \emph{arXiv preprint arXiv:2403.08629}, 2024{\natexlab{b}}.

\bibitem[Kulkarni et~al.(2024)Kulkarni, Rempe, Genova, Kundu, Johnson, Fouhey, and Guibas]{kulkarni2024nifty}
Nilesh Kulkarni, Davis Rempe, Kyle Genova, Abhijit Kundu, Justin Johnson, David Fouhey, and Leonidas Guibas.
\newblock Nifty: Neural object interaction fields for guided human motion synthesis.
\newblock In \emph{Proceedings of the IEEE/CVF Conference on Computer Vision and Pattern Recognition}, pages 947--957, 2024.

\bibitem[Li et~al.(2023{\natexlab{a}})Li, Clegg, Mottaghi, Wu, Puig, and Liu]{li2023controllable}
Jiaman Li, Alexander Clegg, Roozbeh Mottaghi, Jiajun Wu, Xavier Puig, and C~Karen Liu.
\newblock Controllable human-object interaction synthesis.
\newblock \emph{arXiv preprint arXiv:2312.03913}, 2023{\natexlab{a}}.

\bibitem[Li et~al.(2023{\natexlab{b}})Li, Wu, and Liu]{li2023object}
Jiaman Li, Jiajun Wu, and C~Karen Liu.
\newblock Object motion guided human motion synthesis.
\newblock \emph{ACM Transactions on Graphics (TOG)}, 42\penalty0 (6):\penalty0 1--11, 2023{\natexlab{b}}.

\bibitem[Liang et~al.(2024)Liang, Zhang, Li, Yu, and Xu]{liang2024intergen}
Han Liang, Wenqian Zhang, Wenxuan Li, Jingyi Yu, and Lan Xu.
\newblock Intergen: Diffusion-based multi-human motion generation under complex interactions.
\newblock \emph{International Journal of Computer Vision}, pages 1--21, 2024.

\bibitem[Lin(2005)]{lin2005recall}
C Lin.
\newblock Recall-oriented understudy for gisting evaluation (rouge).
\newblock \emph{Retrieved August}, 20:\penalty0 2005, 2005.

\bibitem[Loper et~al.(2023)Loper, Mahmood, Romero, Pons-Moll, and Black]{loper2023smpl}
Matthew Loper, Naureen Mahmood, Javier Romero, Gerard Pons-Moll, and Michael~J Black.
\newblock Smpl: A skinned multi-person linear model.
\newblock In \emph{Seminal Graphics Papers: Pushing the Boundaries, Volume 2}, pages 851--866. 2023.

\bibitem[Papinesi(2002)]{papinesi2002bleu}
K Papinesi.
\newblock Bleu: A method for automatic evaluation of machine translation.
\newblock In \emph{Proc. 40th Actual Meeting of the Association for Computational Linguistics (ACL), 2002}, pages 311--318, 2002.

\bibitem[Peng et~al.(2023)Peng, Xie, Wu, Jampani, Sun, and Jiang]{peng2023hoi}
Xiaogang Peng, Yiming Xie, Zizhao Wu, Varun Jampani, Deqing Sun, and Huaizu Jiang.
\newblock Hoi-diff: Text-driven synthesis of 3d human-object interactions using diffusion models.
\newblock \emph{arXiv preprint arXiv:2312.06553}, 2023.

\bibitem[Prokudin et~al.(2019)Prokudin, Lassner, and Romero]{prokudin2019efficient}
Sergey Prokudin, Christoph Lassner, and Javier Romero.
\newblock Efficient learning on point clouds with basis point sets.
\newblock In \emph{Proceedings of the IEEE/CVF international conference on computer vision}, pages 4332--4341, 2019.

\bibitem[Radford et~al.(2021)Radford, Kim, Hallacy, Ramesh, Goh, Agarwal, Sastry, Askell, Mishkin, Clark, et~al.]{clip}
Alec Radford, Jong~Wook Kim, Chris Hallacy, Aditya Ramesh, Gabriel Goh, Sandhini Agarwal, Girish Sastry, Amanda Askell, Pamela Mishkin, Jack Clark, et~al.
\newblock Learning transferable visual models from natural language supervision.
\newblock In \emph{International conference on machine learning}, pages 8748--8763. PMLR, 2021.

\bibitem[Roumeliotis and Tselikas(2023)]{roumeliotis2023chatgpt}
Konstantinos~I Roumeliotis and Nikolaos~D Tselikas.
\newblock Chatgpt and open-ai models: A preliminary review.
\newblock \emph{Future Internet}, 15\penalty0 (6):\penalty0 192, 2023.

\bibitem[Shafir et~al.(2023)Shafir, Tevet, Kapon, and Bermano]{mdmprior}
Yonatan Shafir, Guy Tevet, Roy Kapon, and Amit~H Bermano.
\newblock Human motion diffusion as a generative prior.
\newblock \emph{arXiv preprint arXiv:2303.01418}, 2023.

\bibitem[Tevet et~al.(2022)Tevet, Raab, Gordon, Shafir, Cohen-or, and Bermano]{mdm2022human}
Guy Tevet, Sigal Raab, Brian Gordon, Yoni Shafir, Daniel Cohen-or, and Amit~Haim Bermano.
\newblock Human motion diffusion model.
\newblock In \emph{The Eleventh International Conference on Learning Representations}, 2022.

\bibitem[Touvron et~al.(2023)Touvron, Lavril, Izacard, Martinet, Lachaux, Lacroix, Rozi{\`e}re, Goyal, Hambro, Azhar, et~al.]{touvron2023llama}
Hugo Touvron, Thibaut Lavril, Gautier Izacard, Xavier Martinet, Marie-Anne Lachaux, Timoth{\'e}e Lacroix, Baptiste Rozi{\`e}re, Naman Goyal, Eric Hambro, Faisal Azhar, et~al.
\newblock Llama: Open and efficient foundation language models.
\newblock \emph{arXiv preprint arXiv:2302.13971}, 2023.

\bibitem[Wan et~al.(2023)Wan, Dou, Komura, Wang, Jayaraman, and Liu]{wan2023tlcontrol}
Weilin Wan, Zhiyang Dou, Taku Komura, Wenping Wang, Dinesh Jayaraman, and Lingjie Liu.
\newblock Tlcontrol: Trajectory and language control for human motion synthesis.
\newblock \emph{arXiv preprint arXiv:2311.17135}, 2023.

\bibitem[Wang et~al.(2021)Wang, Xu, Xu, Liu, and Wang]{wang2021synthesizing}
Jiashun Wang, Huazhe Xu, Jingwei Xu, Sifei Liu, and Xiaolong Wang.
\newblock Synthesizing long-term 3d human motion and interaction in 3d scenes.
\newblock In \emph{Proceedings of the IEEE/CVF Conference on Computer Vision and Pattern Recognition}, pages 9401--9411, 2021.

\bibitem[Wang et~al.(2022)Wang, Chen, Liu, Zhu, Liang, and Huang]{wang2022humanise}
Zan Wang, Yixin Chen, Tengyu Liu, Yixin Zhu, Wei Liang, and Siyuan Huang.
\newblock Humanise: Language-conditioned human motion generation in 3d scenes.
\newblock \emph{Advances in Neural Information Processing Systems}, 35:\penalty0 14959--14971, 2022.

\bibitem[Wang et~al.(2023)Wang, Wang, Lin, and Dai]{wang2023intercontrol}
Zhenzhi Wang, Jingbo Wang, Dahua Lin, and Bo Dai.
\newblock Intercontrol: Generate human motion interactions by controlling every joint.
\newblock \emph{arXiv preprint arXiv:2311.15864}, 2023.

\bibitem[Wang et~al.(2024)Wang, Chen, Jia, Li, Zhang, Zhang, Liu, Zhu, Liang, and Huang]{wang2024move}
Zan Wang, Yixin Chen, Baoxiong Jia, Puhao Li, Jinlu Zhang, Jingze Zhang, Tengyu Liu, Yixin Zhu, Wei Liang, and Siyuan Huang.
\newblock Move as you say interact as you can: Language-guided human motion generation with scene affordance.
\newblock In \emph{Proceedings of the IEEE/CVF CVPR}, pages 433--444, 2024.

\bibitem[Wu et~al.(2024)Wu, Shi, Huang, Yu, Xu, and Wang]{wu2024thor}
Qianyang Wu, Ye Shi, Xiaoshui Huang, Jingyi Yu, Lan Xu, and Jingya Wang.
\newblock Thor: Text to human-object interaction diffusion via relation intervention.
\newblock \emph{arXiv preprint arXiv:2403.11208}, 2024.

\bibitem[Xiao et~al.(2023)Xiao, Wang, Wang, Cao, Zhang, Dai, Lin, and Pang]{xiao2023unified}
Zeqi Xiao, Tai Wang, Jingbo Wang, Jinkun Cao, Wenwei Zhang, Bo Dai, Dahua Lin, and Jiangmiao Pang.
\newblock Unified human-scene interaction via prompted chain-of-contacts.
\newblock \emph{arXiv preprint arXiv:2309.07918}, 2023.

\bibitem[Xie et~al.(2023)Xie, Jampani, Zhong, Sun, and Jiang]{xie2023omnicontrol}
Yiming Xie, Varun Jampani, Lei Zhong, Deqing Sun, and Huaizu Jiang.
\newblock Omnicontrol: Control any joint at any time for human motion generation.
\newblock \emph{arXiv preprint arXiv:2310.08580}, 2023.

\bibitem[Xu et~al.(2024{\natexlab{a}})Xu, Lv, Yan, Jin, Wu, Xu, Liu, Zhou, Rao, Sheng, et~al.]{xu2024inter}
Liang Xu, Xintao Lv, Yichao Yan, Xin Jin, Shuwen Wu, Congsheng Xu, Yifan Liu, Yizhou Zhou, Fengyun Rao, Xingdong Sheng, et~al.
\newblock Inter-x: Towards versatile human-human interaction analysis.
\newblock In \emph{Proceedings of the IEEE/CVF Conference on Computer Vision and Pattern Recognition}, pages 22260--22271, 2024{\natexlab{a}}.

\bibitem[Xu et~al.(2024{\natexlab{b}})Xu, Zhou, Yan, Jin, Zhu, Rao, Yang, and Zeng]{xu2024regennet}
Liang Xu, Yizhou Zhou, Yichao Yan, Xin Jin, Wenhan Zhu, Fengyun Rao, Xiaokang Yang, and Wenjun Zeng.
\newblock Regennet: Towards human action-reaction synthesis.
\newblock In \emph{Proceedings of the IEEE/CVF Conference on Computer Vision and Pattern Recognition}, pages 1759--1769, 2024{\natexlab{b}}.

\bibitem[Xu et~al.(2023)Xu, Li, Wang, and Gui]{xu2023interdiff}
Sirui Xu, Zhengyuan Li, Yu-Xiong Wang, and Liang-Yan Gui.
\newblock Interdiff: Generating 3d human-object interactions with physics-informed diffusion.
\newblock In \emph{Proceedings of the IEEE/CVF International Conference on Computer Vision}, pages 14928--14940, 2023.

\bibitem[Yi et~al.(2025)Yi, Thies, Black, Peng, and Rempe]{yi2025generating}
Hongwei Yi, Justus Thies, Michael~J Black, Xue~Bin Peng, and Davis Rempe.
\newblock Generating human interaction motions in scenes with text control.
\newblock In \emph{ECCV}, pages 246--263. Springer, 2025.

\bibitem[Zhang et~al.(2025)Zhang, Zhang, Dong, Zang, and Wang]{zhang2025long}
Beichen Zhang, Pan Zhang, Xiaoyi Dong, Yuhang Zang, and Jiaqi Wang.
\newblock Long-clip: Unlocking the long-text capability of clip.
\newblock In \emph{ECCV}, pages 310--325. Springer, 2025.

\bibitem[Zhang et~al.(2023{\natexlab{a}})Zhang, Luo, Yang, Xu, Wu, Shi, Yu, Xu, and Wang]{zhang2023neuraldome}
Juze Zhang, Haimin Luo, Hongdi Yang, Xinru Xu, Qianyang Wu, Ye Shi, Jingyi Yu, Lan Xu, and Jingya Wang.
\newblock Neuraldome: A neural modeling pipeline on multi-view human-object interactions.
\newblock In \emph{Proceedings of the IEEE/CVF Conference on Computer Vision and Pattern Recognition}, pages 8834--8845, 2023{\natexlab{a}}.

\bibitem[Zhang et~al.(2023{\natexlab{b}})Zhang, Rao, and Agrawala]{zhang2023adding}
Lvmin Zhang, Anyi Rao, and Maneesh Agrawala.
\newblock Adding conditional control to text-to-image diffusion models.
\newblock In \emph{Proceedings of the IEEE/CVF International Conference on Computer Vision}, pages 3836--3847, 2023{\natexlab{b}}.

\bibitem[Zhang et~al.(2022)Zhang, Cai, Pan, Hong, Guo, Yang, and Liu]{motiondiffuse}
Mingyuan Zhang, Zhongang Cai, Liang Pan, Fangzhou Hong, Xinying Guo, Lei Yang, and Ziwei Liu.
\newblock Motiondiffuse: Text-driven human motion generation with diffusion model.
\newblock \emph{arXiv preprint arXiv:2208.15001}, 2022.

\bibitem[Zhang et~al.(2024)Zhang, Cai, Pan, Hong, Guo, Yang, and Liu]{zhang2024motiondiffuse}
Mingyuan Zhang, Zhongang Cai, Liang Pan, Fangzhou Hong, Xinying Guo, Lei Yang, and Ziwei Liu.
\newblock Motiondiffuse: Text-driven human motion generation with diffusion model.
\newblock \emph{IEEE Transactions on Pattern Analysis and Machine Intelligence}, 2024.

\bibitem[Zhao et~al.(2024)Zhao, Zhang, Du, Shan, Wang, Yu, Wang, and Xu]{Zhao_2024_CVPR}
Chengfeng Zhao, Juze Zhang, Jiashen Du, Ziwei Shan, Junye Wang, Jingyi Yu, Jingya Wang, and Lan Xu.
\newblock I'm hoi: Inertia-aware monocular capture of 3d human-object interactions.
\newblock In \emph{Proceedings of the IEEE/CVF Conference on Computer Vision and Pattern Recognition (CVPR)}, pages 729--741, 2024.

\end{thebibliography}
}

\clearpage
\setcounter{page}{1}
\maketitlesupplementary
We present additional details of implement and LLM annotation in Sec~\ref{sec:add}. Additional experiments are provided in Sec~\ref{sec:sup-exp}, including the user study, generation results after first stage, more ablation studies and visualization results.

\section{Additional Details.}
\label{sec:add}
\subsection{Implement details.}
We implemented our model using Pytorch with training on NVIDIA A40 GPU. Batch size $B$ = 16. We use AdamW optimizer, and the learning rate is $0.0002$. 
The dimension of human motion $D$ = 263, followed MDM~\cite{mdm2022human}, and point cloud is downsampled to $N$ = 1024. The text feature dimension $F_{text}$ encoded from CLIP~\cite{clip} and LONGCLIP~\cite{zhang2025long} is 512. 
We initialize the parameters of both the original MDM and Motion ControlNet from the pretrained MDM weights, freezing the parameters of the original MDM during training. The multi-head attention in transformer for extracting latent features is configured with $n_{head} = 4$.

For the dataset implementation, IMHD$^2$ and HoDome are previously released for motion capture tasks. We extend these two human-object interaction datasets for the contextual generation task. Specifically, we divide the original long motion sequences into shorter clips of length $L=100$, with a frame rate of 30 FPS. Each clip is annotated with a corresponding text description with our LLM annotation module. For IMHD$^2$, we establish a new benchmark by dividing the dataset into training and testing sets based on different subjects, with 1000 clips for training and 800 clips for testing. For HoDome, we manually remove parts of the dataset with poor object reconstruction quality. The remaining data serves as unseen samples to evaluate the generalization capabilities of the model.
\subsection{LLM annotation details.}
We further provide the details of LLM Annotation in Tab.~\ref{tab:LLM} and Tab.~\ref{tab:LLM2}. We also provide several generated textual descriptions in Fig.~\ref{fig:more}, illustrating the texts are well-aligned with the corresponding generated motions, enhancing the interpretability and coherence of the results.

\section{Additional Experiments}
\label{sec:sup-exp}
\subsection{User study.}
We conducte a user study to further evaluate our approach. Specifically, we generated 30 sequences for each method on FullManipulationBody and ask ten participants to rate the results of each interaction, the score range from 1 to 5, higher is better. The evaluation was based on two dimensions: contact score, which measures the feasibility and stability of contact, and physical rationality, which evaluates whether the motions are physically plausible (e.g., avoiding foot slippage or unnatural distortions). Notably, human actions can occasionally exhibit slight distortions even when the contact score is relatively high. Our algorithm addresses this trade-off by incorporating multi-level semantic and geometric constraints and loss guidance to maintain a balance.
The results of the study are presented in Tab.~\ref{tab:omomo-user}. Our method outperforms others, achieving a higher contact score and greater rationality, with the help of the integration of semantic and geometric information in our design.

\begin{table}[t!]
\centering
\caption{User study for human motion generation result on FullBodyManipulation.}
\label{tab:omomo-user}
\vspace{-1ex}
\resizebox{\linewidth}{!}{ 
\begin{tabular}{c|l|ccc}
\rowcolor[HTML]{D9E1F4} 
\hline
\multicolumn{1}{l|}{} & Model & Contact Score & Rationality & Total Score\\ \hline
\multirow{2}{*}{w/o text} 
& SceneDiff ~\cite{scenediff}  & 1.53 & 2.12 & 1.83 \\ 
& OMOMO~\cite{li2023object}   & 4.11 & 3.63 & 3.84 \\ \hline
\multirow{3}{*}{w GT text} 
& CHOIS  \cite{li2023controllable}  & 3.39 & 3.57 & 3.39 \\
& MDM-PC~\cite{mdm2022human}   & 3.07 & 3.64 & 3.35 \\ 
& AffordMotion~\cite{wang2024move}  & 1.81 & 2.71& 2.23 \\ \hline
\multirow{1}{*}{w Gen text} 
& SemGeoMo & \textbf{4.61} & \textbf{4.28} & \textbf{4.45} \\ \hline
\end{tabular}}
\vspace{-2ex}
\end{table}

\begin{table*}[ht!]
\caption{Ablation study on different baselines.}
\resizebox{\linewidth}{!}{ 
\centering
\begin{tabular}{c|cccccccccccc}
\rowcolor[HTML]{D9E1F4} 
\hline
joint constrain & HandJPE$\downarrow$                                & MPJPE   $\downarrow$                               & $C_{prec} \uparrow$                       & $C_{rec}\uparrow$                       & $C_{acc}\uparrow$ & c\%$\uparrow$    & F1 $\uparrow$   & FID $\downarrow$ & R-score $\uparrow$   & Diversity $\uparrow$ & FS $\downarrow$  \\ \hline
OMOMO  &33.18 & 18.06 & 0.77 & 0.71 & 0.74 & 0.61 & 0.75 & 1.98 & 0.38 & 8.99 & 0.50\\\hline
OMOMO-Text   &30.52 & 17.53 & 0.78 & 0.72 & 0.76 & 0.62 & 0.76 & 1.26 & 0.43 & 9.18 & 0.48 \\\hline
OMOMO-Loss   &29.16 & 17.01 & 0.79 & 0.73 & 0.77 & 0.64 & 0.76 & 1.96 & 0.36 & 8.96 & 0.47 \\\hline
\end{tabular}
}
\label{tab:sup-omomo}
\end{table*}
\subsection{Additional experiment results.}
We further provide the metrics from the first stage SemGeo Hierarchical Guidance Generation. 
We observe that during interactions, the hand is the primary joint in contact with the object. Therefore, we predict the hand joint and incorporate hand-joint specific guidance in the second stage to enhance the motion generation. To evaluate generation performance in the first stage, we use Hand\_JPE, which measures the difference between the predicted hand joint position and the ground truth, and Cosine Similarity measures the quality of the generated affordance map. Experiment result is shown in Tab.~\ref{tab:hand}. 
The objects in FullBodyManipulation~\cite{li2023object} are larger compared to other datasets, resulting in more diverse affordance areas, which reduces the similarity between the generated results and the ground truth. And the interactions in IMHD$^2$~\cite{Zhao_2024_CVPR} are more challenging, making the generation process harder and leading to relatively lower accuracy in joint predictions. 

\subsection{Ablation studies on other baseline.}
We further conduct experiments on OMOMO~\cite{li2023object} to validate the effectiveness of incorporating textual information and the impact of loss guidance, denoted as OMOMO-Text and OMOMO-Loss. After integrating language information, the FID score significantly improves, indicating that the generation results are more rational and coherent. 

\begin{table}[t!]
\centering
\caption{Joint generation result from SemGeo Hierarchical Guidance Generation.}
\vspace{-1ex}
\resizebox{\linewidth}{!}{ 
\begin{tabular}{l|cccc}
\rowcolor[HTML]{D9E1F4} 
\hline
& Left\_JPE & Right\_JPE & Hand\_JPE & Sim \\\hline
 FullBodyManipulation~\cite{li2023object}   & 28.84 & 27.96 & 28.40 & 0.27\\ 
 Behave~\cite{bhatnagar2022behave}   & 28.94& 27.06 & 28.00 & 0.15\\
 IMHD$^2$~\cite{Zhao_2024_CVPR}    & 30.49 & 40.26 & 35.38 & 0.52 \\
\hline
\end{tabular}}
\vspace{-2ex}
\label{tab:hand}
\end{table}

\begin{table}[t]
\small
  \begin{center}
  \caption{\textbf{Detailed prompting example for LLM Annotation.}}
  \label{tab:LLM} 
    \begin{tabular}{p{235pt}}
    \toprule
    Detailed prompting example for LLM Annotation. \\
    \midrule
    Instructions: You are an expert on the interaction between 3D human motion and object. A person will interact with a object, give me a sentence that how the person will interact with this object based on following information.
  \\  \midrule
$[$start of Given Information$]$\\
Coordinate System: \\
The coordinate system of the 3D scene includes x, y, and z-axes. The person moves on the XOZ plane, and the positive y-axis represents height.\\
Target Object and category:\\ 
The category of the object is CLASS. The size of the CLASS is SIZE.\\
The interaction with this object will last approximately T seconds.
The object center: CENTER. \\
Possible actions list: ACTION\_LIST = [`face', `flip', `grab', `grasp', `hold', `kick', `lift', `move', `pick', `place', `push', `pull', `put', `release', `rotate', `set', `slide', `swing', `tilt', `turn', `sit', `bend', `shake', `wave', `drag'] \\
$[$End of Given Information$]$\\
    \bottomrule
    \end{tabular}
  \end{center}
\end{table}

\begin{table}[t]
\small
  \begin{center}
  \caption{\textbf{Detailed prompting example for fine-grained LLM Annotation.}}
  \label{tab:LLM2} 
    \begin{tabular}{p{235pt}}
    \toprule
    Detailed prompting example for LLM Annotation. \\
    \midrule
    Instructions: A person lift the plasticbox, rotate the plasticbox, and set it back down.
  \\  \midrule
You are an expert on the interaction between 3D human motion and object. Given the instruction, give me a sentence that how the person will interact with this object in detailed, including the arm and leg movement in each 3s, make each sentence just include key action.\\
$[$start of Given Information$]$\\
Coordinate System: \\
The coordinate system of the 3D scene includes x, y, and z-axes. The person moves on the XOZ plane, and the positive y-axis represents height.\\
Target Object and category:\\ 
The category of the object is CLASS. The size of the CLASS is SIZE.\\
The object center with hand contact information in total LAST\_TIME \\
At T, object center is CENTER, [no hand contact, single contact hand(left/right), both hand in contact] at position POS.\\
$[$End of Given Information$]$\\
\\

$[$Start of Rule$]$\\
Divide the total movement into three step.\\
Inference how their arms and legs move.\\
Inference the hand-object interaction direction,  chosen from:``left", ``right", ``top", ``bottom",``left-top", ``left-bottom", ``right-top", and ``right-bottom".\\
Make sure each sentence includes key action.\\
$[$End of Rule$]$\\
\\
$[$Start of Example$]$\\
A person lifts the white chair, rotates the white chair, and puts down the white chair.\\
The fine-grained result:\\
First, the person faces the back of the white chair, grasps it with both hands from the left-bottom and right-bottom sides, bending slightly at the knees as both arms lift the chair off the ground.  
Next, maintaining grip, the person rotates the white chair with both hands,lifting the chair slightly higher.  
Finally, the person puts down the white chair, with the right arm pushing from the right-top and the left arm steadying from the left-top, as both legs move forward to reposition. \\ 
$[$End of Example$]$\\
    \bottomrule
    \end{tabular}
  \end{center}
\end{table}

\subsection{Additional visualization results.}
\noindent\textbf{Comparison with ground truth.}
We provide a sample demonstrating that, given a sequential point cloud, the potential interactions can be diverse. The ground truth interaction is not the only valid solution. While our generated motion may differ from the ground truth, it remains rational and contextually appropriate.

\begin{figure}
    \centering
\includegraphics[width=0.95\linewidth]{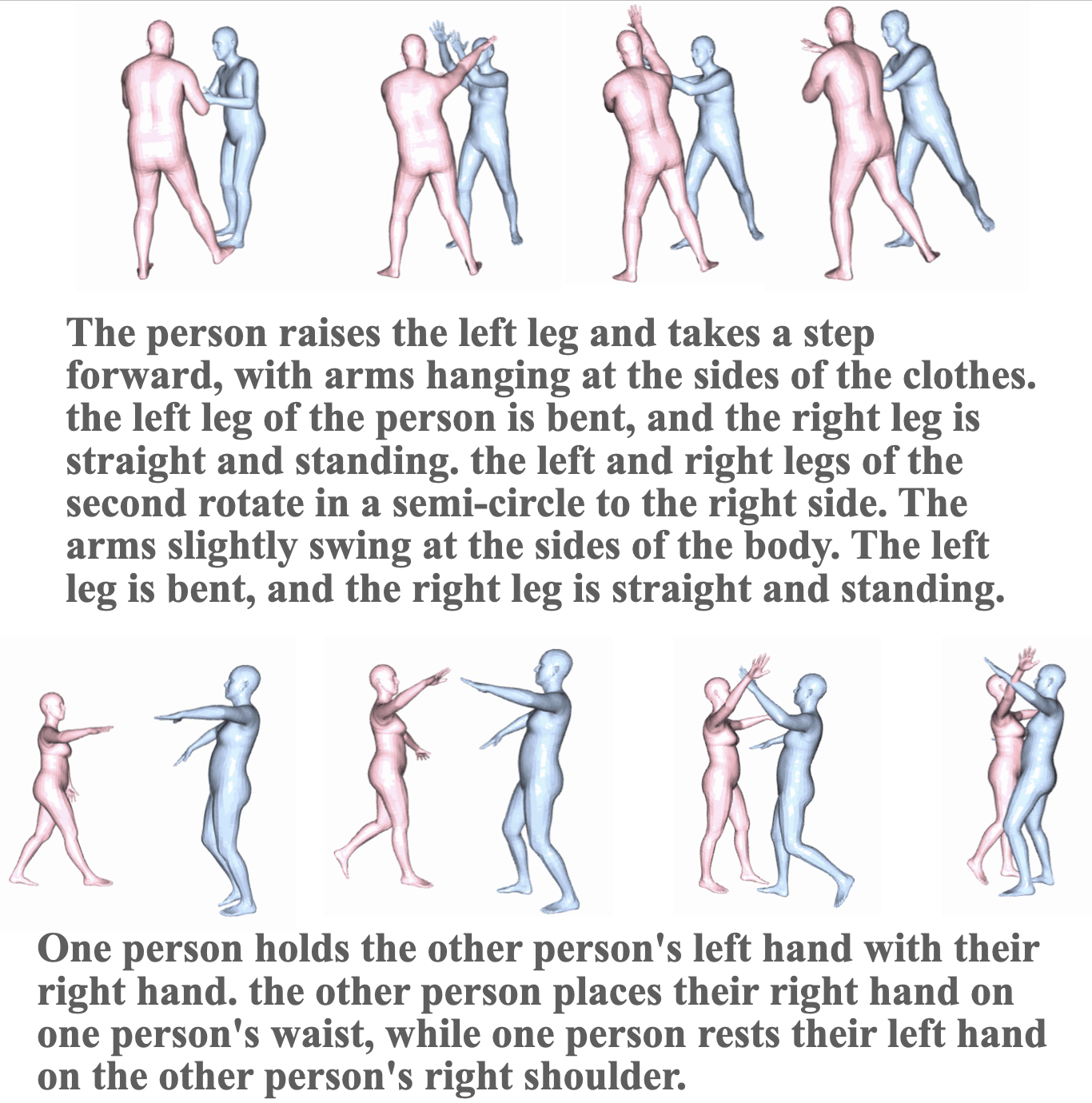}
    \caption{Extension on human-human generation.}
    \label{fig:more-human}
\end{figure}

\begin{figure*}
    \centering
\includegraphics[width=0.98\linewidth]{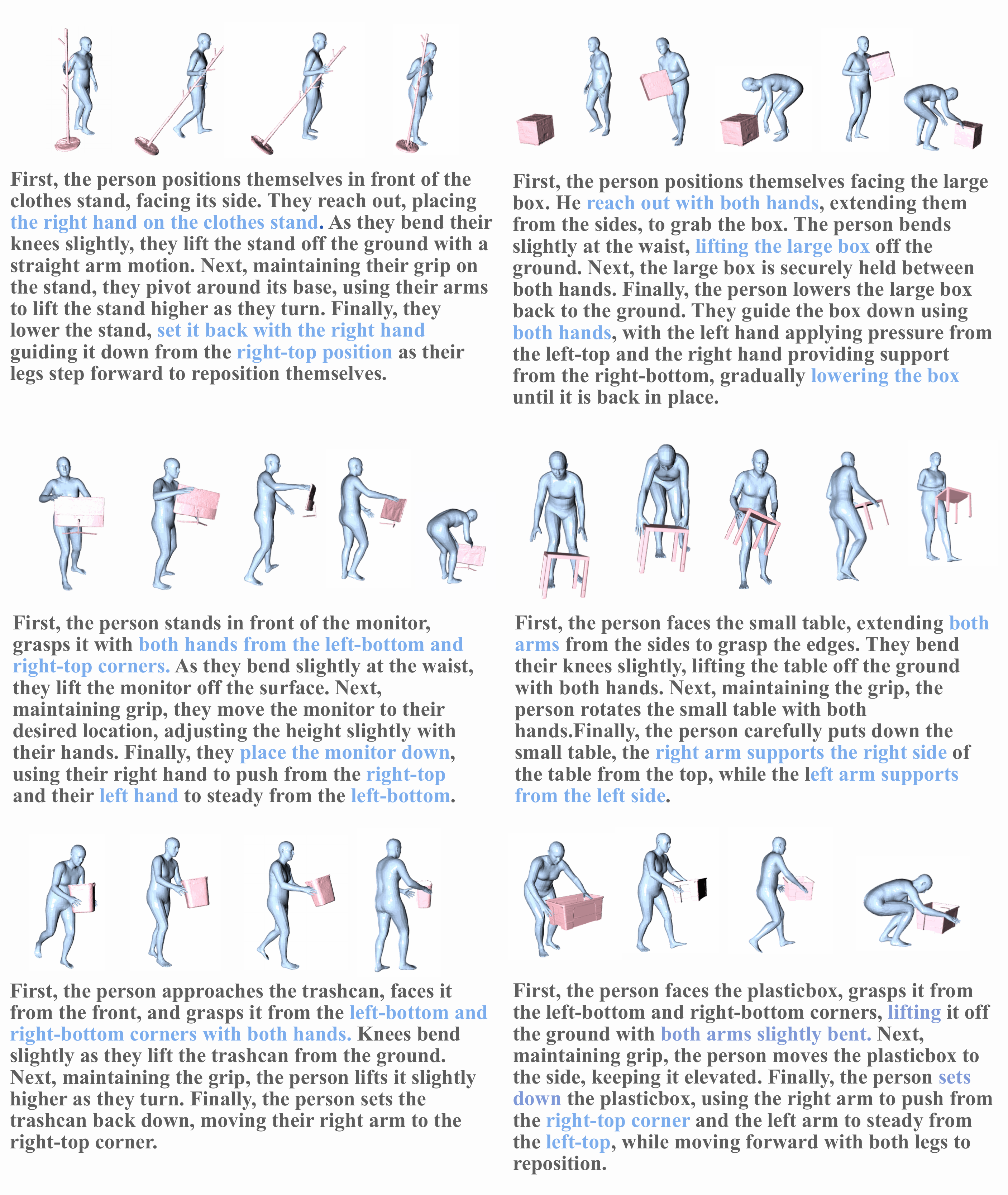}
    \caption{The generation result is aligned with the generated fine-grained text.}
    \label{fig:more}
    \vspace{4ex}
\end{figure*}

\noindent\textbf{More generation results.}
We present additional generation results in Figure.~\ref{fig:more} and include a detailed comparison and demonstration of the results in our accompanying video.
\noindent\textbf{More extension results.}
We present extension generation results in Figure.~\ref{fig:more-human} and demonstration of the results in our accompanying video.

\begin{figure}[ht!]
    \centering
\includegraphics[width=0.85\linewidth]{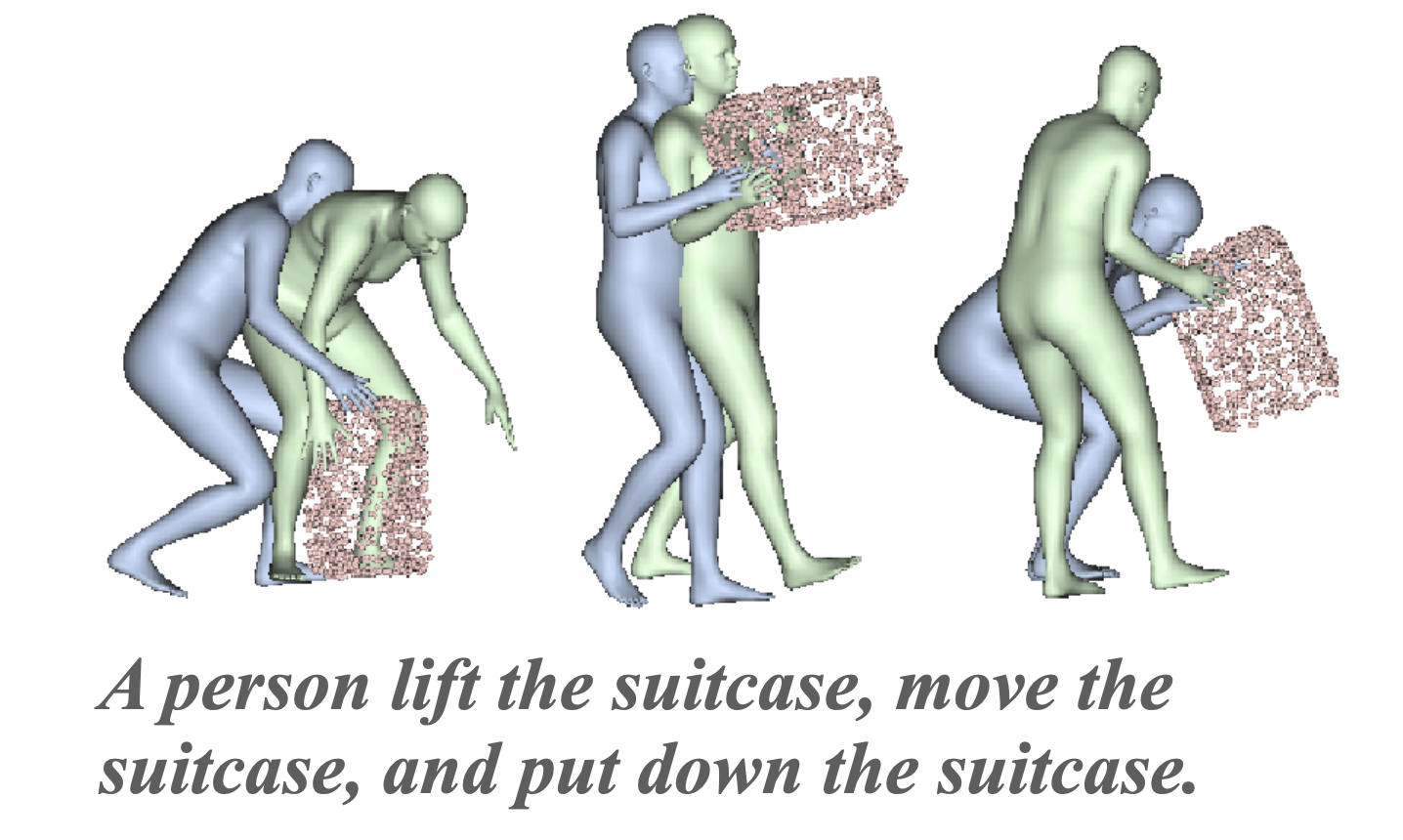}
    \caption{Visualization compared with ground truth, the potential interaction can be diverse. The person in green is the ground truth and the blue is our generation result.}
    \label{fig:gt_gen}
\end{figure}

\end{document}